\documentclass[accepted]{uai2023} % after acceptance, for a revised
\usepackage[american]{babel}
\usepackage{natbib} % has a nice set of citation styles and commands
\usepackage{mathtools} % amsmath with fixes and additions
\usepackage{booktabs} % commands to create good-looking tables
\usepackage{tikz} % nice language for creating drawings and diagrams

\title{Interpretable Differencing of Machine Learning Models}

\author[1]{\href{mailto:<swagatam.haldar@ibm.com>?Subject=Your UAI 2023 paper}{Swagatam~Haldar}{}}
\author[1]{Diptikalyan~Saha}
\author[2]{Dennis~Wei}
\author[3]{Rahul~Nair}
\author[3]{Elizabeth~M.~Daly}
\affil[1]{%
    IBM Research\\
    Bangalore, India
}
\affil[2]{%
    IBM Research\\
    Yorktown Heights, New York, USA\\
}
\affil[3]{%
    IBM Research\\
    Dublin, Ireland
  }

\usepackage{hyperref}
\usepackage{booktabs}
\usepackage{multirow}
\usepackage{amsmath}
\usepackage{amsfonts}
\usepackage{amssymb}
\usepackage{amsthm}
\usepackage[linesnumbered,ruled]{algorithm2e}
\usepackage{balance}

\usepackage{rotating}  % to avoid using \small, rotate tables

\usepackage{tikz}

\usetikzlibrary{shapes,arrows,positioning} 
\usetikzlibrary{shapes.geometric}
\tikzset{
    >=stealth',
    punkt/.style={
           rectangle,
           draw=black,
           text width=6.5em,
           minimum height=3em,
           text centered},
     box/.style={
      rectangle, 
      draw=black, 
      minimum width=2em,
      minimum height=2em,
      text centered
  },
    pil/.style={
           ->,
           shorten <=2pt,
           shorten >=2pt,},
    revpil/.style={
           <-,
           shorten <=2pt,
           shorten >=2pt,}
}

\newcommand{\eat}[1]{}
\DeclareMathOperator*{\argmin}{arg\,min}
\DeclareMathOperator*{\imp}{imp}
\DeclareMathOperator*{\lab}{label}
\DeclareMathOperator*{\leaves}{leaves}
\DeclareMathOperator*{\pc}{pc}
\newcommand{\Dtest}{\mathcal{D}_{\mathrm{test}}}
\newcommand{\Dtrain}{\mathcal{D}_{\mathrm{train}}}
\newcommand{\Ttrue}{\mathcal{T}_{\mathrm{true}}}
\newcommand{\Tpred}{\mathcal{T}_{\mathrm{pred}}}

\begin{document}
\maketitle

\begin{abstract}
Understanding the differences between machine learning (ML) models is of interest in scenarios ranging from choosing amongst a set of competing models, to updating a deployed model with new training data. In these cases, we wish to go beyond differences in overall metrics such as accuracy to identify \emph{where} in the feature space do the differences occur. We formalize this problem of model \emph{differencing} as one of predicting a dissimilarity function of two ML models' outputs, subject to the representation of the differences being human-interpretable. Our solution is to learn a \emph{Joint Surrogate Tree} (JST), which is composed of two conjoined decision tree surrogates for the two models. A JST provides an intuitive representation of differences
and places the changes in the context of the models' decision logic. Context is important as it helps users to map differences to an underlying mental model of an AI system. We also propose a refinement procedure to increase the precision of a JST. 
We demonstrate, through an empirical evaluation, that such contextual differencing is concise and can be achieved with no loss in fidelity over naive approaches.
\end{abstract}

%%%%%%%%%%%%%%%%%%%%%%%%%%%%%%%%%%%%%%%%%%
%%%%%%%%%%%%%%%%%%%%%%%%%%%%%%%%%%%%%%%%%%

\section{INTRODUCTION}
\label{sec: intro}

% Motivation
At various stages of the AI model lifecycle, data scientists make decisions regarding which model to use. For instance, they may choose from a range of pre-built models, select from a list of candidate models generated from automated tools like AutoML, or simply update a model based on new training data to incorporate distributional changes. In these settings, the choice of a model is preceded by an evaluation that typically focuses on accuracy and other metrics, instead of how it differs from other models.   

% Summary statement of problem
We address the problem of model \emph{differencing}. Given two models for the same task and a dataset, we seek to learn where in the feature space the models' predicted outcomes differ. Our objective is to provide accurate and interpretable mechanisms to uncover these differences.

The comparison is helpful in several scenarios. In a \emph{model marketplace}, multiple pre-built models for the same task need to be compared. The models usually are black-box and possibly trained on different sets of data drawn from the same distribution. During \emph{model selection}, a data scientist trains multiple models and needs to select one model for deployment. In this setting, the models are white-box and typically trained on the same training data. For \emph{model change}, where a model is retrained with updated training data with a goal towards model improvement, the data scientist needs to understand changes in the model beyond accuracy metrics. Finally, \emph{decision pipelines consisting of logic and ML models} occur in business contexts where a combination of business logic and the output of ML models work together for a final output. Changes might occur either due to model retraining or adjustments in business logic which can impact the behavior of the overall pipeline.

% Summary of our proposal
In this work we address the problem of interpretable model differencing as follows. First, we formulate the problem as one of predicting the values of a {dissimilarity function} of the two models' outputs. We focus herein on $0$-$1$ dissimilarity for two 
%(possibly multi-class) 
classifiers, where $0$ means ``same output'' and $1$ means ``different'', so that prediction quality can be quantified by any binary classification metric such as precision and recall.
Second, we propose a method that learns a \emph{Joint Surrogate Tree} (JST), composed of two conjoined decision tree surrogates to jointly approximate the two models. The root and lower branches of the conjoined decision trees are common to both models, while higher branches (farther from root) may be specific to one model. %JSTs thus align the two surrogate models to allow easier comparison, while also avoiding the challenge of directly modelling the differences, which tend to be fragmented. We present a visualization of JSTs as an intuitive representation of model differences. 
A JST thus accomplishes two tasks at once: it provides interpretable surrogates for the two models while also aligning the surrogates for easier comparison and identification of differences. These aspects are encapsulated in a visualization of JSTs that we present.
%The setting we explore is similar to \cite{nair2021changed} which uses rule-based surrogates. Our approach avoids the two limitations of their method around biases in inducing rules and remove the one-to-one correspondence restriction. 
Third, a refinement procedure is used to grow the surrogates in selected regions, improving the {precision} of the dissimilarity prediction. 

% % Blurb about mental models.
Our design of jointly learning surrogates is motivated by the need to place model differences in the context of the overall decision logic. This can aid users who may already %tend to 
have a mental model of (individual) AI systems, either for debugging \citep{kulesza2012tell} or to understand errors \citep{bansal2019beyond}.

% Contribution statement
The main contributions of the paper are (a) a quantitative formulation of the problem of model differencing, and (b) algorithms to learn and refine conjoined decision tree surrogates to approximate two models simultaneously. A detailed evaluation of the method is presented on several benchmark datasets, showing more accurate or more concise representation of model differences, compared to baselines.

%%%%%%%%%%%%%%%%%%%%%%%%%%%%%%%%%%%%%%%%%%%
%%%%%%%%%%%%%%%%%%%%%%%%%%%%%%%%%%%%%%%%%%%

\section{RELATED WORKS}

Our work touches upon several active areas of research which we summarize based on key pertinent themes.

\paragraph{Surrogate models and model refinement}
One mechanism to lend interpretability to machine learning models is through surrogates, i.e., simpler human-readable models that mimic a complex model \citep{bucila2006model,ba2014do,hinton2015distilling,lopez-paz2016unifying}. Most relevant to this paper are works that use a decision tree as the surrogate \citep{TREPAN,bastani2017interpretability,frosst2017distilling}. \citet{bastani2017interpretability} showed that interpretable surrogate decision trees extracted from a black-box ML model allowed users to predict the same outcome as the original ML model. %The authors showed that by comparing the decision tree surrogates of two medical providers, differences in how diagnoses were reported could be identified.
\citet{freitas2014comprehensible} also discusses interpretability and usefulness of using decision trees as surrogates. None of these works however have considered jointly approximating two black-box models.

\paragraph{Decision tree generation with additional objectives}  \citet{chen2019robust} showed that decision tree generation is not robust and slight changes in the root node can result in a very different tree structure. %Work presented in 
\citet{chen2019robust, andriushchenko2019provably} focus on improving robustness when generating the decision tree while \citet{moshkovitz2021connecting} prioritises both robustness and interpretability. \citet{aghaei2019learning} use mixed-integer optimization to take fairness into account in the decision tree generation. However, none of these solutions consider the task of comparing two decision trees. 

%However, when considering the task of comparing two models, the additional challenge of ensuring the tree structure is relatively comparable becomes important. 

\paragraph{Predicting disagreement or shift}
Prior work has focused on identifying statistically whether models have significantly changed \citep{Bu2019_modelchangedetection,Geng2019_ChangeDetection,harel2014}, but not on where they have changed. \citet{cito2021explaining} present a model-agnostic rule-induction algorithm to produce interpretable rules capturing instances that are mispredicted with respect to their ground truth.

\paragraph{Comparing models} The ``distill-and-compare'' approach of \citet{tan2018distill} uses generalized additive models (GAMs) and fits one GAM to a black-box model and a second GAM to ground truth outcomes. While differences between the GAMs are studied to uncover insights, there is only one black-box model. \citet{Demsar2018DetectingCD} study concept drift by determining feature contributions to a model and observing the changes in contributions over time. Similarly, \citet{duckworth2021using} investigated changes in feature importance rankings pre- and post-COVID. This approach however does not localize changes to regions of the feature space. %could be used to identify data drift, however while it can be used to signal drift has occurred it does not provide insights on which populations in the data the predictions have changed and how those predictions have changed.
\citet{chouldechova2017fairer} compare models in terms of fairness metrics and identify groups in the data where two models have maximum disparity.
Prior work by \citet{nair2021changed}, which is most similar to our own, uses rule-based surrogates for two models and derives rules for where the models behave similarly. %The output of 
%Their method is more qualitative as it does not evaluate the accuracy of the rules in predicting model similarities or differences. %mappings. 
%In contrast to our solution, 
Their method biases the learning of the second surrogate based on inputs from the first model, a step they call grounding, and imposes a one-to-one mapping between rules in the two surrogates. This is a strict condition that may not hold in practice. Additionally, their method does not evaluate the accuracy of resulting rules in predicting model similarities or differences. Our approach addresses these limitations.

%%%%%%%%%%%%%%%%%%%%%%%%%%%%%%%%%%%%%%%%%%%
%%%%%%%%%%%%%%%%%%%%%%%%%%%%%%%%%%%%%%%%%%%

\section{PROBLEM STATEMENT AND PRELIMINARIES}
\label{sec:prob}

%\paragraph{Problem statement} 
We are given two predictive models $M_1, M_2: \mathcal{X} \to \mathcal{Y}$ mapping a feature space $\mathcal{X} \subset \mathbb{R}^d$ to an output space $\mathcal{Y}$, as well as a \emph{dissimilarity function} $D: \mathcal{Y} \times \mathcal{Y} \to \mathbb{R}_+$ (where $\mathbb{R}_+$ means the non-negative reals including zero) for comparing the outputs of the two models. Our goal is to obtain a \emph{difference model} (``diff-model'' for short), $\hat{D}: \mathcal{X} \to \mathbb{R}_+$, that predicts the dissimilarity $D(M_1(x), M_2(x))$ well while also being interpretable. To construct $\hat{D}$, we assume access to a dataset $X \in \mathbb{R}^{n\times d}$ consisting of $n$ samples drawn i.i.d.~from a probability distribution $P$ over $\mathcal{X}$. This dataset does not have to have ground truth labels, in contrast to supervised learning, since supervision is provided by the models $M_1, M_2$. Prediction quality is measured by the expectation $\mathbb{E}[L(\hat{D}(X), D(M_1(X), M_2(X))]$ of one or more metrics $L: \mathbb{R}_+ \times \mathbb{R}_+ \to \mathbb{R}_+$ comparing $\hat{D}$ to $D$, where the expectation is with respect to $P$. In practice, these expectations are approximated empirically using a test set.

In this work, we focus on classification models $M_1$ and $M_2$, so that $\mathcal{Y}$ is a finite set, and $0$-$1$ dissimilarity $D(M_1(x), M_2(x)) = 1$ if $M_1(x) \neq M_2(x)$ and $D(M_1(x), M_2(x)) = 0$ otherwise. 
% Extension to regression models is discussed in Section~\ref{sec:concl}. 
Accordingly, the predictions $\hat{D}(x)$ are also binary-valued and any binary classification metrics $L$ may be used for evaluation. Herein we use precision, recall, and F1-score (described in Section~\ref{sec:expt}). 

%To ensure that the difference model $\hat{D}$ is interpretable, we restrict attention to decision trees as the model class and
We use decision trees as the basis for our Joint Surrogate Tree solution. To ensure interpretability, the height (also referred to as maximum depth) is constrained to a small value (e.g.~$6$ in our experiments). Below we define notation and terminology related to decision trees for later use.

\paragraph{Decision Tree} A decision tree is a binary tree $T=(V_{dt},E_{dt})$ with a node set $V_{dt}$, a root node $r \in V_{dt}$ and a directed set of edges $E_{dt} \subset V_{dt} \times V_{dt}$. Each internal node $v \in V_{dt}$ contains a split condition $s(v) := f(v) < t(v)$ containing a predicate on feature $f(v) \in [d]$ (where $[d]$ is the  shorthand for $\{1\ldots d\}$), and a threshold $t(v) \in \mathbb{R}$, and  two children $v_T$ and $v_F$. The edges $(v,v_T)$ and $(v,v_F)$ are annotated with edge conditions $f(v) <  t(v)$ and $f(v) \ge  t(v)$, respectively. Each leaf node $v$ contains a label $\lab(v) \in \mathcal{Y}$. All leaf nodes of a tree rooted at $r$ are denoted as $\leaves(r)$. Given a node $v$, path-condition of $v$ (denoted as $\pc(v)$) is defined as the conjunction of all edge conditions from $r$ to $v$. At a given node $v \in V_{dt}$, we denote by $X_v, y_v$ %and $X_v[f]$ as 
the subset of samples that satisfy the $pc(v)$ and their labels, and we use $X_v[f]$ to denote the set of values for the feature $f\in [d]$. Without loss of generality, $s(v)$ is formed by minimizing function $H$, for all features and their values. We express  the split condition at node $v$ as $s(v) = c(X_v,y_v)$ and the minimum objective value (impurity) by $\imp(X_v,y_v)$:
\begin{align} 
c(X_v,y_v)&=\argmin_{\{f\in[d],\,t\in X_v[f]\}} H(f,t,X_v,y_v)\\
\imp(X_v,y_v)&=\min_{\{f\in[d],\,t\in X_v[f]\}} H(f,t,X_v,y_v) \label{eq:imp}
\end{align}

For example, $H$ can be instantiated as the weighted sum of  entropy values of left and right split~\citep{RQ}. We now describe two baseline approaches to the problem before presenting our proposed algorithm in Section~\ref{sec:algo}.

\paragraph{Direct difference modelling} Given the above problem statement, a natural %naive 
way to predict the dissimilarity function $D$ is to let $\hat{D}$ be a single ML model, in our case a decision tree for interpretability, and train it to classify between $D=0$ (models $M_1, M_2$ having the same output) and $D=1$ (different output). We call this the \emph{direct} approach. The main drawback of direct differencing is that even when using an interpretable decision tree, it does not capture the differences between the two models in the context of their human-interpretable decision processes, i.e., where in the decision logic of the models do the differences occur. 

\paragraph{Surrogate modelling} Another natural %naive 
way to model the dissimilarity is to separately build a decision tree surrogate $\hat{M}_i$ for each input model $M_i$, $i=1,2$, using the outputs of $M_i$ on the input samples $X$ for training the surrogate. Then we predict $\hat{D}(x) = 1$ if $\hat{M}_1(x) \neq \hat{M}_2(x)$ and $\hat{D}(x) = 0$ otherwise. We call this the \emph{separate surrogate} approach. Its drawback is that the two decision tree surrogates are not aligned, making it cumbersome %and un-intuitive 
for human comparison. In Section~\ref{sec:expt}, we show that the manifestation of this drawback is the large number of rules (see next paragraph) needed to describe all the regions where the two surrogates differ. 

\paragraph{Diff rules as output} 
We use \emph{diff rules} as an interpretable representation of model differences for both direct and surrogate tree-based diff models. A diff rule is a conjunction of conditions on individual features that, when satisfied at a point $x$, yields the prediction $\hat{D}(x) = 1$. Corresponding to each diff rule is a \emph{diff region}, the set of $x$'s that satisfy the rule. A \emph{diff ruleset} $\mathcal{R}$ is a set of diff rules such that if $x$ satisfies any rule in the set, we predict $\hat{D}(x) = 1$. For a direct decision tree model, the diff rules are given by the path conditions of the $\hat{D}(x) = 1$ leaves. For surrogate models $\hat{M}_1, \hat{M}_2$, the diff rules are conjunctions of path conditions for pairs of intersecting leaves where $\hat{M}_1(x) \neq \hat{M}_2(x)$.

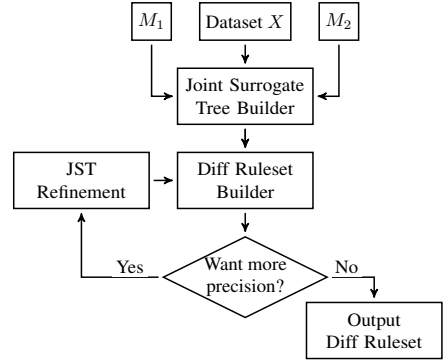
\begin{figure}[t]
    \centering
    \resizebox{0.7\columnwidth}{!}{
\begin{tikzpicture}[node distance=0.5cm, auto,thick]
  % steps
  \node[punkt, align=center] (jst) {Joint Surrogate \\ Tree Builder};
  \node[punkt, below=of jst] (diff) {Diff Ruleset\\ Builder}
     edge[revpil] (jst);
  \node[draw,diamond,aspect=2,align=center, inner sep=0.1mm, below=of diff](inter){Want more\\ precision?};
    %Still\\ Interpretable?};
     edge[revpil] (diff);
  %\node[punkt, below left=of inter](sample){Sample Generation};
  %\node[draw,diamond, aspect=2, align=center, xshift=-1em, yshift=-3em, inner sep=0.1mm, below left=of inter](fid){Low\\ Fidelity?};
  \node[punkt, below right=of inter,yshift=0.8em](outd){Output\\ Diff Ruleset};
  \node[punkt, left=of diff](ref){JST\\ Refinement};
  \node[box,above=of jst](in_s){Dataset $X$};
  \node[box, left=of in_s](M1){$M_1$};
  \node[box, right=of in_s](M2){$M_2$};
 
  \draw[pil] (in_s) edge (jst)
     (jst) edge (diff)
     (diff) edge (inter);
 
 \draw[pil] (inter.west) -| (ref.south)
     node[pos=0.2,fill=white,inner sep=0.2em, anchor=south]{Yes};
 \draw[pil] (inter.east) -| (outd.north)
     node[pos=0.2,fill=white,inner sep=0.2em, anchor=south]{No};
 
 \draw[pil](M1.south) |- (jst.west);
 \draw[pil](M2.south) |- (jst.east);
 
 %\draw[pil] (sample) to (fid);
 \draw[pil] (ref) to (diff);
 
% \draw[pil] (fid.west) -- node [anchor=south]{Yes} ++(-1,0) |- (ref.west);
 
 %\draw[pil] (fid.east) -| (outd.south)
  %   node[pos=0.1,fill=white,inner sep=0.2em]{No};
 
 \end{tikzpicture}
        }
\caption{Method Overview}
\label{fig:method}
\end{figure}

\section{PROPOSED ALGORITHM}
\label{sec:algo}

We propose a technique called IMD, which shows the differences between two ML models by constructing a novel representation called a \emph{Joint Surrogate Tree} or JST. A JST is composed of two conjoined decision tree surrogates that jointly approximate the two models, intuitively capturing similarities and differences between them.
It overcomes the drawbacks of the direct and separate surrogate approaches mentioned in Section~\ref{sec:prob}: it avoids the non-smoothness of direct difference modelling, aligns and promotes similarity between surrogates for the two models, and shows differences in the context of each model's decision logic. Our method has a single hyperparameter, tree depth, which controls the trade-off between accuracy and interpretability. 

IMD performs two steps as shown in Figure~\ref{fig:method}. In the first step, IMD builds a JST for models $M_1, M_2$ using data samples $X$, and then extracts diff regions from the JST. Interpretability is ensured by restricting the height of the JST. The IMD algorithm treats $M_1, M_2$ as black boxes and can handle any pair of classification models. It is also easy to implement as it requires a simple modification to popular greedy decision tree algorithms. 

The second (optional) step, discussed at the end of  Section~\ref{sec:algo:refine}, refines the JST by identifying diff regions where the two decision tree surrogates within the JST differ but the original models do not agree with the surrogates on their predictions. The refinement process aims to increase the fidelity of the surrogates in the diff regions, thereby generating more precise diff regions where the true models also differ.  

\subsection{Joint Surrogate Tree}
\label{section:joint-surrogate-tree}

\paragraph{Representation}

\begin{figure}[t]
    % \centering
\resizebox{\columnwidth}{!}{

\begin{tikzpicture}
   \definecolor{m1}{HTML} {ffb6c1ff};
   \definecolor{m2}{HTML}{ffa07aff};
   \definecolor{l1}{HTML}{e0ffffff};
   \definecolor{l0}{HTML}{faebd7ff};
   \node[anchor=south west,inner sep=0] (jst) at (0,0) {\includegraphics[width=\textwidth]{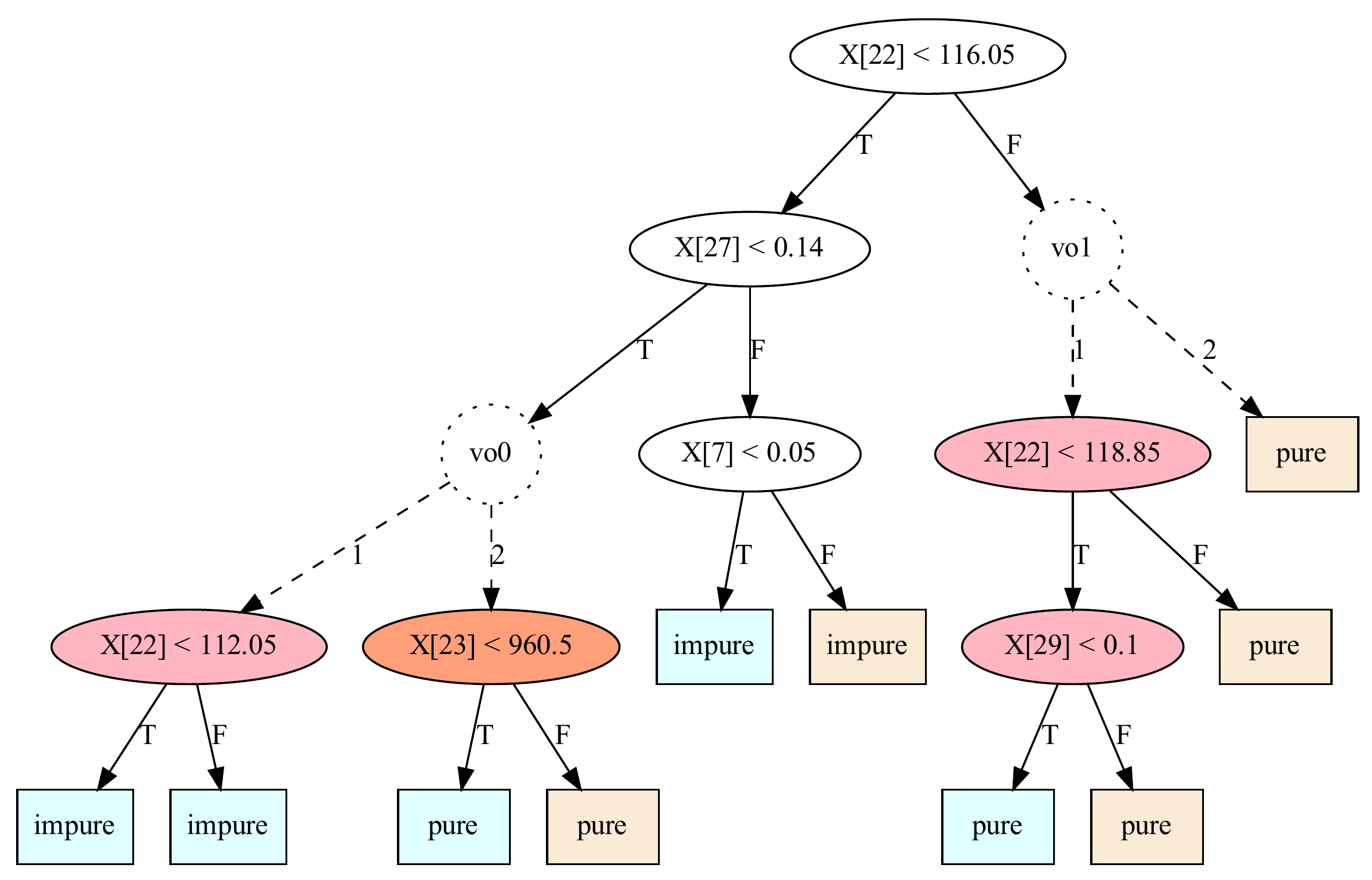}};
   \begin{scope}[x={(jst.south east)},y={(jst.north west)}]
        \node[ellipse,draw,scale=1.3] (e) at (0.1,0.9) {Common};
        \node[ellipse,draw=black,fill=m1,scale=1.3](m1) at (0.1, 0.8) {$M_1$ - LR};
        \node[ellipse,draw=black,fill=m2,scale=1.3](m2) at (0.1, 0.7) {$M_2$ - RF};
        \node[circle,draw=black,dashed,align=center] at (0.1, 0.55) {Diverging\\ nodes};
        \node[rectangle,draw=black,fill=l1,scale=1.35] at (0.25,0.9){Label 1};
        \node[rectangle,draw=black,fill=l0,scale=1.35] at (0.25,0.8){Label 0};
        
        %\draw[help lines,xstep=.1,ystep=.1] (0,0) grid (1,1);
        %\foreach \x in {0,1,...,9} { \node [anchor=north] at (\x/10,0) {0.\x}; }
        %\foreach \y in {0,1,...,9} { \node [anchor=east] at (0,\y/10) {0.\y}; }
    \end{scope}
\end{tikzpicture}

        }
\caption{A JST for the Breast Cancer (\textit{bc}) dataset.}
\label{fig:jst}
\end{figure}
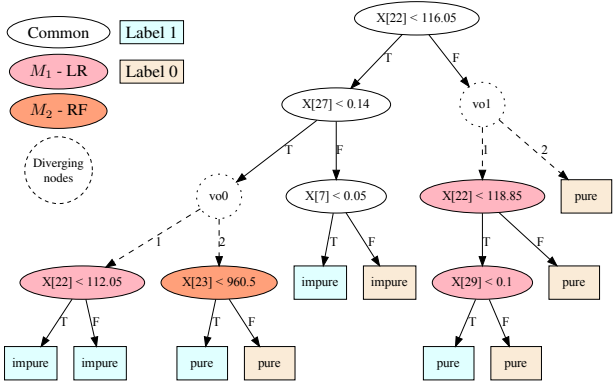

%Our key model comparison output JST is shown in 
Figure~\ref{fig:jst} shows an example of a JST for Logistic Regression and Random Forest models on the Breast cancer dataset~\citep{UCIRepo} (feature names are omitted to save space). 
%As mentioned, 
The JST consists of two conjoined decision tree surrogates for the two models. The white oval nodes of the JST are shared decision nodes where both surrogates use the same split conditions. We refer to the subtree consisting of white nodes as the common prefix tree.
In contrast, the colored nodes represent separate decision nodes, pink for surrogate $\hat{M}_1$ corresponding to $M_1$, and orange for surrogate $\hat{M}_2$ for $M_2$. The rectangular nodes correspond to the leaves, and are colored differently to represent class labels --- cyan for label 1, and beige for label 0. The leaves are marked as pure/impure depending on whether all the samples falling there have the same label or not.

The JST intuitively captures diff regions, i.e., local regions of feature space where the two input models diverge, and also groups them into a two-level hierarchy. As with any surrogate-based diff model, we have $\hat{D}(x) = 1$ if and only if the constituent decision tree surrogates disagree, $\hat{M}_1(x) \neq \hat{M}_2(x)$. 
Thus, diff regions can be identified by first focusing on an \emph{or-node} (the dotted circle nodes in Figure~\ref{fig:jst} where the surrogates diverge) and then enumerating pairs of leaves under it with different labels.

For example, considering the rightmost or-node $v_{o1}$ in Figure~\ref{fig:jst}, with path condition $X[22] \ge 116.05$, $\hat{M}_2$ classifies all the samples to label $0$ whereas $\hat{M}_1$ classifies to label $1$ in the region $X[22] < 118.85 \wedge X[29] < 0.1$. Therefore the diff region is $118.85 > X[22] \ge 116.05 \wedge X[29] < 0.1$. 
While in this case $v_{o1}$ yields a single diff region, in general multiple diff regions could be grouped under a single or-node, resulting in a hierarchy. 
By processing all the or-nodes of the JST, one obtains all diff-regions.  

Formally, $JST = (V=V_{dt} \cup V_o, E=E_{dt}\cup E_o)$. $V_{dt}$ is a set of decision nodes similar to decision trees (oval shaped in figure) with each outgoing edge $\in E_{dt}$ (solid arrows) representing True or False decisions as in a regular decision tree. $V_o$ are the set of or-nodes (circular nodes) representing the diverging points where the decision trees no longer share the same split conditions. Each child of $v_o\in V_o$ is denoted as $v_o^{i}$, $i = 1, 2$, with dashed edges $(v_o, v_o^{i}) \in E_o$.  Each $v_o^{i}$ represents the root of an individual surrogate decision sub-tree for model $i$. The height of a JST is the maximum number of decision edges (solid edges) in any root-to-leaf path. 

%The formal definition of the diff regions is as follows. 
Formally, a diff region is defined by the non-empty intersection of path-conditions of differently labelled leaves $l_1, l_2$ from two decision sub-trees rooted at the same or-node $v_o$. The collection of all diff regions specifies the diff ruleset:
\begin{align}
\mathcal{R} = &\left\{\pc(l_1)\wedge \pc(l_2) : l_i\in \leaves(v^i_o), \; i=1,2, \right.\nonumber\\
&\left. \quad \lab(l_1)\neq \lab(l_2), \; v_o\in V_o \right\}.\label{eq:diff-region}
\end{align}

\paragraph{Construction} 
The objective of JST construction is %therefore 
two-fold:
(a) \emph{Maximize comparability}: To achieve maximal sharing of split conditions between the two decision tree surrogates, and
(b) \emph{Interpretability}: Achieve the above objective under the constraint of interpretability. We have chosen the height of the JST as the interpretability constraint.

The construction of a JST corresponding to the inputs $M_1, M_2, X$ starts with evaluating  $y_1=M_1(X)$ and $y_2=M_2(X)$. Starting from the root, at each internal node $v \in V_{dt}$, with inputs $(X_v,y_{1v}=M_1(X_v),y_{2v}=M_2(X_v))$ filtered by the node's path condition, the key choice is whether to create a joint decision node or an or-node for the surrogates to \emph{diverge}. The choice of node type signifies whether the two surrogates will continue to share their split conditions or not. %Note that, 
Once divergence happens at an or-node, the two sub-trees rooted at the or-node %both child decision trees 
do not share any split nodes thereafter. 
Below we present a general condition for divergence and a simplified one implemented in our experiments. 

In general, a divergence condition should compare the cost of a joint split to that of separate splits for the two models. In the context of greedy decision tree algorithms considered in this work, the comparison is between the sum of impurities for the best possible common split,
%In the case of non-divergence, the following objective function returns the common split condition: 
%
%\begin{equation}
%\mathop{\mathrm{argmin}}_{\{f\in[d],\,t\in X_v[f]\}} H(f,t,X_v,y_{1v}) + H(f,t,X_v,y_{2v})
%\end{equation}
\begin{multline}\label{eq:impJoint}
    \imp(X_v, y_{1v}, y_{2v}) =\\ \min_{\{f\in[d],\,t\in X_v[f]\}} H(f,t,X_v,y_{1v}) + H(f,t,X_v,y_{2v}),
\end{multline}
and the impurities $\imp(X_v, y_{1v})$, $\imp(X_v, y_{2v})$ \eqref{eq:imp} for the best separate splits. One condition for divergence is 
\begin{equation}\label{eq:divCond1}
    \imp(X_v, y_{1v}) + \imp(X_v, y_{2v}) \leq \alpha \imp(X_v, y_{1v}, y_{2v})
\end{equation}
for some $\alpha \leq 1$. %$\alpha \in [0, 1]$. 
The choice $\alpha = 1$ always results in divergence and thus reduces to the separate surrogate approach in Section~\ref{sec:prob}. This happens because the left-hand side of \eqref{eq:divCond1} corresponds to separately minimizing the two terms in \eqref{eq:impJoint}, hence ensuring that \eqref{eq:divCond1} is true. As $\alpha$ decreases, joint splits are  favored. %For $\alpha = 0$, divergence occurs only when both impurities are zero, implying that all children of the two splits are pure. %The $imp$ value of zero implies that both the children are pure.
For $\alpha < 0$, divergence essentially never occurs.\footnote{If $\imp(X_v, y_{1v}, y_{2v}) = 0$, then $\imp(X_v, y_{1v}) = \imp(X_v, y_{2v}) = 0$ also and the same $(f, t)$ pair minimizes all three impurities. Hence divergence has no effect.}

%The second condition amounts to evaluating the impurity function for all splits corresponding to both $(X_v,y_{1v})$ and $(X_v,y_{2v})$ and  
For this work, we choose to heavily bias the algorithm toward joint splits and greater interpretability of the resulting JST. In this case, we use the simplified condition 
\begin{equation}\label{eq:divCond2}
\imp(X_v,y_{1v}) = 0  \vee \imp(X_v,y_{2v}) = 0,
\end{equation}
%\vspace{0.5cm}
which results in divergence if at least one of the minimum impurity values is zero. The advantage of \eqref{eq:divCond2} over \eqref{eq:divCond1} is that the minimization in \eqref{eq:impJoint} to compute $\imp(X_v, y_{1v}, y_{2v})$ can be done lazily, only if \eqref{eq:divCond2} is not satisfied. If condition \eqref{eq:divCond2} is met, we create an or-node, two or-edges, and grow individual surrogate trees from that point onward. Figure~\ref{fig:jst} shows 1 instance (node $v_{o0}$) where \eqref{eq:divCond2} is met. A special case of \eqref{eq:divCond2} occurs %A trivial condition for divergence is  
when at least one of $y_{1v}, y_{2v}$ contains only one label, i.e., it is already pure without splitting. The node $v_{o1}$ in Figure~\ref{fig:jst} shows one such case. 

The JST construction ends if pure leaf nodes are found or the height of the JST has reached a pre-defined hyper-parameter value $k$. 

\paragraph{JST Refinement}
\label{sec:algo:refine}

We now present an iterative process for refinement aimed at increasing precision of diff regions.% \eqref{eq:diff-region}. 

For each leaf $l_i$ contributing to a diff region \eqref{eq:diff-region}, if its samples (satisfying $\pc(l_i)$) have more than one label as given by the model $M_i$ being approximated (the leaf is impure), we can further split it into two leaf nodes. This refines the decision tree surrogates only in the diff regions and not at all impure leaves. Next, diff regions are recomputed with the resulting leaf nodes. This process can continue for a pre-defined number of steps or until some budget is met. Every such iteration increases the tree depth by $1$ (but not uniformly) and improves the fidelity of the individual sub-tree rooted at an or-node. 

%%%%%%%%%%%%%%%%%%%%%%%%%%%%%%%%%%%%%%%%%%%
%%%%%%%%%%%%%%%%%%%%%%%%%%%%%%%%%%%%%%%%%%%

\section{EXPERIMENTAL RESULTS}
\label{sec:expt}

% @Swagatam
We report experimental results comparing the proposed IMD technique to learning separate surrogates for the two models (Section~\ref{sec:expt:sep}), and to direct difference modelling and the prior work of \citet{nair2021changed} (Section~\ref{sec:expt:alt}). The effect of refinement is demonstrated in Section~\ref{sec:expt:ref}. The following paragraphs describe the setup of the experiments.

\paragraph{Datasets} We have used 13 publicly available~\citep{UCIRepo,OpenML_as_a_whole,alcala2011keel} tabular classification datasets, including both binary and multiclass classification tasks. As preprocessing steps, we dropped duplicate instances occurring in the original data, and one-hot encoded categorical features.

\paragraph{Models} We split each dataset in the standard 70/30 ratio, and trained an array of machine learning models --- Decision Tree Classifier (DT), Random Forest Classifier (RF), K-Neighbours Classifier (KN), Logistic Regression (LR), Gradient Boosting (GB), Multi-Layered Perceptron (MLP), and Gaussian Naive Bayes (GNB). For some models, multiple instances were trained with different parameter values. We have used the Scikit-learn~\citep{scikit-learn} implementations for training. Once trained, we did not do any performance tuning of the models, and used them as black boxes (through the $\tt{predict()}$ method only) for subsequent analyses.
The dataset and model details including test set accuracies are reported in the supplementary material (SM). %~\ref{appdx: bench} and ~\ref{appdx: models}.

\paragraph{Set Up} We have selected two pairs of models per dataset corresponding to the largest and smallest differences in accuracy on the test set (indicated as \textit{max $M_1$-$M_2$} and \textit{min $M_1$-$M_2$} in Table~\ref{tab:ablations-with-deltas-trimmed-for-uai}). This ensures we compare models with contrasting
predictive performance, as well as models that achieve similar accuracy. For fitting and evaluating diff models, including our IMD approach as well as baselines, we split the available dataset $X$ (without labels) in a 70/30 ratio into $\Dtrain$ and $\Dtest$. This split is not and does not have to be the same as the train/test splits for training and evaluating the underlying models.
We perform 5 train/test splits and report in the main paper the mean of the following metrics across the 5 runs, with standard deviation values in the SM.

\paragraph{Metrics} To measure how accurately we capture the true regions of disagreement between models $M_1$ and $M_2$, %using the rules, 
we use the following metrics. Given a test set $\Dtest$, we have a subset of \emph{true diff samples}:
\[ \Ttrue = \{ x \in \Dtest \,|\, M_1(x) \neq M_2(x)\}, \]
and the predicted diff samples by the diff model $\hat{D}(x)$:
\[ \Tpred = \{ x \in \Dtest \,|\, \hat{D}(x) = 1
\}.
\]
Recall that in the case where we have extracted a diff ruleset $\mathcal{R}$ for $\hat{D}(x)$, $x \in \Tpred$ if there exists a rule $r \in \mathcal{R}$ that is satisfied by $x$.

\noindent{\bf Precision (Pr)} is the ratio $\frac{| \Ttrue\, \cap \, \Tpred|}{|\Tpred|}$, measuring the fraction of predicted diff samples that are true diff samples on the test set $\Dtest$.

\noindent{\bf Recall (Re)} is the ratio $\frac{|\Ttrue \, \cap \, \Tpred|}{|\Ttrue|}$, measuring the fraction of true diff samples in $\Dtest$ that are correctly predicted.

\noindent{\bf Interpretability} For interpretable diff models for which we have extracted a diff ruleset  $\mathcal{R}$, we measure its interpretability in terms of the number of rules \textbf{(\# r)} in the set, and the number of unique predicates \textbf{(\# p)} summed over all the rules in the set. The choice of the above metrics is motivated by the works of~\citet{lakkaraju2016interpretable,dash2018boolean,letham2015interpretable}.

\subsection{IMD against Separate Surrogates}
\label{sec:expt:sep}

% 
% WITHOUT PREDICATES COLUMN and ADDED DELTA FOR PR/RE/RULES
% 

\begin{table*}[t]
\centering
\caption{Sep. surrogates shows slightly higher recall, but IMD shows comparable performance with much less complexity.
}

\begin{tabular}{cc}
% Table & Same Table \\

        % Table 1

        \begin{tabular}{llcccrccr}
        \toprule
                  &            &     &  \multicolumn{3}{c}{\textbf{Separate Surrogates}} & \multicolumn{3}{c}{\textbf{IMD}} \\
        \cmidrule(r){4-6} \cmidrule(l){7-9}
        \textbf{Dataset}        &  \textbf{$M_1$ vs.~$M_2$} & \textbf{diffs}   &          Pr &      Re &      \#r & Pr &      Re &    \#r \\
        \hline
        \multirow{2}{*}{adult} & max MLP1-GB & 0.20  &           0.96 &  0.88 &    70.0 &             0.96 &  0.88 &  18.0 \\
          & min MLP2-DT2 & 0.08  &           0.45 &  0.29 &   155.4 &             0.46 &  0.16 &  17.4 \\
        \cline{1-9}
        \multirow{2}{*}{bankm} & max MLP2-GB &0.26  &           0.66 &  0.75 &   263.6 &             0.70 &  0.67 &  23.0 \\
                  & min MLP1-GNB & 0.26 &           0.74 &  0.75 &   345.0 &             0.71 &  0.69 &  34.4 \\
        % \cline{1-9}
        % \multirow{2}{*}{banknote} & max KN1-GNB & 0.15  &          0.88 &  0.89 &    32.2 &             0.90 &  0.88 &  13.4 \\
                  % & min LR-DT1 &  0.03  &          0.53 &  0.60 &    30.8 &             0.64 &  0.47 &   7.2 \\
        % \cline{1-9}
        % --\multirow{2}{*}{bc} & max DT1-GNB & 0.05 &           0.38 &  0.46 &    39.2 &             0.44 &  0.40 &   9.6 \\
                  % & min KN2-RF2 & 0.07  &           0.37 &  0.38 &    49.0 &             0.30 &  0.24 &  10.8 \\
        % \cline{1-9}
        % \multirow{2}{*}{diabetes} & max MLP2-GB & 0.22  &           0.42 &  0.45 &   215.8 &             0.40 &  0.28 &  24.2 \\
                  % & min RF1-GNB & 0.18  &           0.39 &  0.43 &   156.0 &             0.31 &  0.34 &  20.8 \\
        \cline{1-9}
        \multirow{2}{*}{eye} & max RF1-GNB &  0.56  &           0.65 &  0.66 &  1054.0 &             0.60 &  0.71 &  36.2 \\
                  & min LR-MLP1 & 0.34  &           0.59 &  0.53 &   781.6 &             0.57 &  0.39 &  28.4 \\
        \cline{1-9}
        \multirow{2}{*}{heloc} & max KN1-RF2 & 0.23 &           0.40 &  0.23 &   373.0 &             0.40 &  0.13 &  15.8 \\
                  & min GB-RF1 & 0.17   &           0.30 &  0.19 &   234.4 &             0.25 &  0.06 &  14.6 \\
        \cline{1-9}
        \multirow{2}{*}{magic} & max RF1-GNB & 0.25 &           0.75 &  0.58 &   362.8 &             0.75 &  0.52 &  25.0 \\
                  & min MLP2-DT2 & 0.11 &           0.43 &  0.36 &   282.6 &             0.42 &  0.17 &  11.0 \\
        % \cline{1-9}
        % \multirow{2}{*}{mushroom} & max KN1-GNB & 0.03  &           0.94 &  0.70 &    28.0 &             0.81 &  0.70 &   5.0 \\
                  % & min RF2-GNB & 0.03  &           0.93 &  0.70 &    42.0 &             0.74 &  0.71 &   8.6 \\
        \cline{1-9}
        \multirow{2}{*}{redwine} & max RF1-KN2 &  0.37  &           0.46 &  0.52 &   627.8 &             0.52 &  0.25 &  29.0 \\
                  & min KN1-GNB & 0.52  &           0.70 &  0.59 &   563.6 &             0.69 &  0.47 &  40.4 \\
        \cline{1-9}
        \multirow{2}{*}{tictactoe} & max LR-GNB & 0.34  &          0.76 &  0.78 &   109.6 &             0.76 &  0.89 &  24.4 \\
                  & min DT2-KN2 & 0.06  &           0.10 &  0.15 &    54.0 &             0.16 &  0.11 &   5.8 \\
        \cline{1-9}
        \multirow{2}{*}{waveform} & max LR-DT1 & 0.18   &           0.45 &  0.52 &   746.0 &             0.49 &  0.27 &  33.2 \\
                  & min MLP1-RF2 & 0.11 &           0.17 &  0.32 &   725.0 &             0.10 &  0.02 &   9.0 \\
        % \cline{1-9}
        % \multirow{2}{*}{whitewine} & max RF1-GNB & 0.53 &          0.64 &  0.59 &   847.2 &             0.63 &  0.56 &  42.6 \\
                  % & min LR-KN2 &  0.48  &           0.56 &  0.33 &   580.0 &             0.55 &  0.35 &  36.6 \\
        \bottomrule
        \end{tabular} &
        
        % Table 2
        
        \begin{tabular}{rrr}
        \toprule
        \multicolumn{3}{c}{\textbf{Sep. $-$ IMD}}   \\
        \cmidrule{1-3}
        $\Delta{\textrm{Pr}}$ &     $\Delta{\textrm{Re}}$ & $\Delta{\textrm{\#r}}$ \\
        
        \hline

            $-$0.00 & $-$0.00 &   $-$52.0 \\
             $+$0.01 & $-$0.13 &  $-$138.0 \\
             \cline{1-3}
             
             $+$0.04 & $-$0.08 &  $-$240.6 \\
            $-$0.03 & $-$0.06 &  $-$310.6 \\
            \cline{1-3}
            
            %  $+$0.01 & $-$0.01 &   $-$18.8 \\
            %  $+$0.11 & $-$0.13 &   $-$23.6 \\
            %  \cline{1-3}
        
            %  $+$0.06 & $-$0.05 &   $-$29.6 \\
            % $-$0.07 & $-$0.14 &   $-$38.2 \\
            % \cline{1-3}
        
            % $-$0.01 & $-$0.17 &  $-$191.6 \\
            % $-$0.08 & $-$0.09 &  $-$135.2 \\
            % \cline{1-3}
        
            $-$0.06 & $+$0.05 & $-$1017.8 \\
            $-$0.02 & $-$0.14 &  $-$753.2 \\
            \cline{1-3}
        
             $+$0.00 & $-$0.10 &  $-$357.2 \\
            $-$0.05 & $-$0.13 &  $-$219.8 \\
            \cline{1-3}
        
             $+$0.00 & $-$0.06 &  $-$337.8 \\
            $-$0.01 & $-$0.18 &  $-$271.6 \\
            \cline{1-3}
        
            % $-$0.13 & $-$0.01 &   $-$23.0 \\
            % $-$0.18 & $+$0.00 &   $-$33.4 \\
            % \cline{1-3}
        
             $+$0.06 & $-$0.27 &  $-$598.8 \\
            $-$0.01 & $-$0.11 &  $-$523.2 \\
            \cline{1-3}
        
            $-$0.00 & $+$0.11 &   $-$85.2 \\
             $+$0.05 & $-$0.04 &   $-$48.2 \\
             \cline{1-3}
        
             $+$0.04 & $-$0.25 &  $-$712.8 \\
            $-$0.07 & $-$0.30 &  $-$716.0 \\
            % \cline{1-3}
        
            % $-$0.01 & $-$0.03 &  $-$804.6 \\
            % $-$0.01 & $+$0.03 &  $-$543.4 \\

        \bottomrule
        \end{tabular} \\
\end{tabular}

\label{tab:ablations-with-deltas-trimmed-for-uai}
\end{table*}

First we study the effect of jointly training surrogates in IMD, which encourages sharing of split nodes, against training separate surrogates for the two models. Since these are both surrogate-based approaches to obtain a diff model $\hat{D}$, we compare the metrics for the \textit{diff rulesets} extracted (as described in Section~\ref{sec:prob}) from the surrogates. IMD extracts diff rulesets from JSTs, while the separate surrogate approach is a special case of IMD corresponding to $\alpha=1$. The height (a.k.a.~maximum depth) of the surrogates is restricted to 6 for both of the approaches. We do not perform the refinement step here as we study it in Section~\ref{sec:expt:ref}.

\noindent{\bf Observations } The metrics are reported for 8 datasets in Table~\ref{tab:ablations-with-deltas-trimmed-for-uai} (full version in Appendix). The differences in Pr, Re, and \# rules are also tabulated for better readability. We also report the fraction of diff samples in $\Dtest$ for each dataset and model pair combination in the ``diffs" column. This value is also the precision of a trivial diff-model ($\hat{D}(x)=1\,\, \forall x$, recall$=1.0$), or any diff-model that predicts \emph{diff} with probability $q$ (recall$=q$), e.g., $q=0.5$ is a random guesser. Clearly, diff prediction quality for both approaches is significantly better than random guessing.

To summarize the table, below we compare the approaches on the basis of average percentage increase or decrease in precision and recall (on going from separate to IMD) across all datasets. We also perform Wilcoxon's signed rank test (as recommended by \citet{benavoli2016should}) to verify the statistical significance of the observed differences.

For precision, we observe a very small drop ($1.55$\% on average) going from separate surrogates to IMD. Wilcoxon's test's $p$-value is $0.269$, implying no significant difference (at level $0.05$) between the approaches. For recall, we observe that IMD has $23.45$\% poorer recall. Wilcoxon's test confirms this difference with a $p$-value of $0.0002$, and a sign test also shows that separate surrogates have higher values of recall for 22 of the 26 benchmarks.

For the interpretability metrics however, IMD is the clear winner looking at the columns corresponding to numbers of rules (\# r) and unique predicates (\# p, in Appendix). If we simply average the numbers of rules and predicates to understand the scale of the difference (with the caveat that different datasets and model pairs have different complexities), the average number of rules for separate and IMD are 337.25 and 20.94, and the average numbers of predicates are 135.41 and 56.10. The corresponding $p$-values are also very low (on the order of $10^{-6}$).

%\noindent{\bf Conclusion } The difference model $\hat{D}$ obtained from separately trained surrogates is somewhat more accurate as evidenced by higher recall, %, and thus F1-score), but it comes with a huge expense in interpretability. The results also affirm the effect of sharing nodes (promoting similarity between surrogates) as much as possible in JST, which localizes differences before divergence.

%\subsection{Comparison among alternate approaches}
\subsection{Comparison with Other Approaches}
\label{sec:expt:alt}

\begin{table}[t]
% \scriptsize
% \small
\centering
\caption{Comparison of F1-scores. The mean ranks ($\downarrow$ the better) highlight that sep. surr., and Direct GB are most accurate, but IMD is close with greater interpretability.}
\begin{tabular}{lccccc}
\toprule
% \hline
          &   & \textbf{Sep.} & \textbf{Direct} & \textbf{Direct} & \textbf{BRCG} \\
          
\textbf{Dataset} & \textbf{IMD} &  \textbf{Surr.} &      \textbf{DT} &        \textbf{GB} &             \textbf{Diff.} \\

% \midrule
\hline
\multirow{2}{*}{adult} &             0.92 &           0.92 &     0.92 &     0.98 &          0.33 \\
          &              0.23 &           0.34 &     0.17 &     0.61 &          0.31 \\
\cline{1-6}
\multirow{2}{*}{bankm} &             0.68 &           0.70 &     0.69 &     0.77 &          0.41 \\
          &              0.70 &           0.75 &     0.68 &     0.82 &          0.41 \\
\cline{1-6}
\multirow{2}{*}{banknote} &             0.89 &           0.89 &     0.83 &     0.94 &          0.27 \\
          &             0.52 &           0.56 &     0.57 &     0.63 &          0.06 \\
\cline{1-6}
\multirow{2}{*}{bc} &             0.39 &           0.41 &     0.17 &     0.00 &          0.10 \\
          &             0.25 &           0.37 &     0.28 &     0.19 &          0.13 \\
\cline{1-6}
\multirow{2}{*}{diabetes} &             0.32 &           0.43 &     0.21 &     0.35 &          0.35 \\
          &             0.32 &           0.41 &     0.09 &     0.22 &          0.30 \\
\cline{1-6}
\multirow{2}{*}{heloc} &             0.19 &           0.29 &     0.03 &     0.14 &          0.37 \\
          &             0.10 &           0.22 &     0.02 &     0.05 &          0.27 \\
\cline{1-6}
\multirow{2}{*}{magic} &             0.62 &           0.65 &     0.63 &     0.78 &          0.40 \\
          &             0.24 &           0.39 &     0.14 &     0.27 &          0.20 \\
\cline{1-6}
\multirow{2}{*}{mushroom} &             0.75 &           0.80 &     0.81 &     0.97 &          0.76 \\
          &             0.72 &           0.80 &     0.81 &     0.97 &          0.74 \\
\cline{1-6}
\multirow{2}{*}{tictactoe} &             0.82 &           0.77 &     0.77 &     0.82 &          0.83 \\
          &             0.12 &           0.12 &     0.00 &     0.09 &          0.00 \\
% \bottomrule
\toprule
\textit{mean rank} &   \textbf{3.278} & \textbf{2.056} &  \textbf{3.694} & \textbf{2.278} & \textbf{3.694} \\

\bottomrule
\end{tabular}

\label{tab:approaches}
\end{table}

In this experiment, we compare the quality of prediction of the true dissimilarity $D$ %the binary \emph{diff-model}
with respect to other baselines. 
The first two baselines are direct approaches (introduced in Section~\ref{sec:prob}) as they relabel the instances as \emph{diff}(``1") or \emph{non-diff}(``0") and directly fit a classification model on the relabeled instances.
Out of a huge number of possible models for this binary classification problem of predicting \textit{diff} or \textit{non-diff}, we choose Decision Tree (with \texttt{max\_depth=6}) to be directly comparable to JST, and Gradient Boosting Classifier (\texttt{max\_depth=6}, rest default settings in Scikit-learn) to provide a more expressive but uninterpretable benchmark.
% We have chosen two direct diff models: Decision Tree (maximum depth restricted to 6) and Gradient Boosting Classifier (maximum depth of 6, other parameters at the default configurations in Scikit-learn) --- which is more expressive, but uninterpretable.
These choices are made to compare the quality of surrogate-based diff regions against directly modelled diff regions, %when the diff-samples are modeled directly, 
and also to understand if we are significantly compromising on  quality by not using a more expressive or uninterpretable model.
As a third baseline, we compare to diff rulesets obtained from Grounded BRCG~\citep{nair2021changed} ruleset surrogates for the two models. 
The surrogate-based approaches from the previous subsection, IMD (without refinement) and separate, are also included for completeness.

\noindent{\bf Observations } We have listed the F1-scores (harmonic mean of precision and recall) in Table~\ref{tab:approaches}, and omitted the $M_1$ vs.~$M_2$ column (same as in Table~\ref{tab:ablations-with-deltas-trimmed-for-uai}) for brevity. Since \textit{BRCG Diff.} applies only to binary classification tasks, we only show it for those.
Note that for IMD and separate surrogates, the precision and recall values are already reported in Table~\ref{tab:ablations-with-deltas-trimmed-for-uai}. For the other methods and datasets, precision, recall, and \# rules (if applicable) are in the Appendix. On average, we observe that IMD achieves a $89.76$\% improvement in F1-score over Direct DT, and $98.52$\% improvement over the BRCG Diff.~approach.\footnote{This is computed by removing the second subrow for tictactoe as F1 score is 0 for both Direct DT and BRCG Diff.~and the jump is infinite. This removal is thus favorable toward them.} On the other hand, we do not observe a large drop in F1-score from the uninterpretable Direct GB to IMD ($-5.87\%$). %\footnote{This %average change 
% is computed by removing the first row for the \textit{bc} dataset as the F1-score for Direct GB is very low ($0.03$) and the relative increase (to $0.34$) is high. It is thus favorable toward Direct GB.} 
Similarly, the precision and recall differences in Section~\ref{sec:expt:sep} combine to give a $-15.26$\% decrease in going from separate surrogates to IMD. 

We report mean ranks in Table~\ref{tab:approaches} and performed Friedman's test following \citet{demsar2006statistical}, which confirms significant differences between the methods with a $p$-value on the order of $0.0006$. Next we perform pairwise comparisons of IMD against the other approaches. 
The $p$-values from Wilcoxon's signed rank test are $0.00025$, $0.043$, $0.043$, and $0.1594$ %for comparisons
against separate, \textit{BRCG Diff.}, \textit{Direct DT}, and \textit{Direct GB} respectively. We pit these against the Holm-corrected thresholds of $0.0125$, $0.017$, $0.025$, $0.05$, and observe that %all of the differences are significant. We emphasize that although separate and direct GB have consistently higher F1-scores than IMD, the size of the differences is small and IMD is considerably more interpretable.
only the first one (IMD vs. separate) is significant for this set of values. However, we emphasize that although separate and Direct GB have consistently higher F1-scores than IMD, the size of the differences is small and IMD is considerably more interpretable.
% \textcolor{blue}{
For the interpretable methods, the average numbers of rules observed for IMD, Direct DT, and BRCG Diff.~are $16.05$, $10.50$, $37.69$ (separate surrogates was already discussed in Section~\ref{sec:expt:sep}).
% }

% \textcolor{blue}{
We present further experiments (in Appendix) varying the depth to understand the accuracy-complexity trade-off for Direct DT, Separate and IMD extensively. While the trade-offs for Direct DT and IMD are competitive, both of them are consistently better than Separate. We also discuss qualitative comparison between Direct DT and IMD which brings out the benefit of IMD in placing the diff rules in the context of the models' decision logic, as already seen in Figure~\ref{fig:jst}.
% }

% \textcolor{red}{Further experiments (in Appendix) on varying depth also corroborate that Direct DT achieves lower F1 scores than IMD on most benchmarks.}

% \textcolor{red}{For the interpretable methods, the average numbers of rules observed for IMD, Direct DT, and BRCG Diff.~are $16.05$, $10.50$, $37.69$ (separate surrogates was already discussed in Section~\ref{sec:expt:sep}). While these numbers are all comparable, the benefit of IMD is that it places the diff rules in the context of the models' decision logic, as seen in Figure~\ref{fig:jst}.}

% We have also seen that BRCG Diff. approach typically achieved higher Recall, but very low Precision scores (not shown in Table~\ref{tab:approaches}) that ultimately brings down its F1.

\subsection{Effect of Refinement}
\label{sec:expt:ref}

\begin{table}
% \small
\centering
\caption{Precision improves on refinement ($\text{IMD}_\text{6+1}$).}

\begin{tabular}{lccc}
\toprule
% \hline
    \textbf{Dataset}     & $\text{\bf IMD}_\text{\bf6}$ & $\text{\bf IMD}_\text{\bf6+1}$ & $\text{\bf IMD}_\text{\bf7}$ \\

\midrule

\multirow{2}{*}{adult} &             0.96 &                   \textbf{0.96} &             0.95 \\
          &              0.46 &                   \textbf{0.59} &             0.53 \\
\cline{1-4}

\multirow{2}{*}{bankm} &            0.70 &                   \textbf{0.78} &             0.77 \\
          &             0.71 &                   \textbf{0.79} &             0.74 \\
\cline{1-4}

% \multirow{2}{*}{banknote}  &             0.90 &                   \textbf{0.90} &             0.89 \\
%           &             0.64 &                   0.67 &             \textbf{0.76} \\
% \cline{1-4}

% \multirow{2}{*}{bc}  &             0.44 &       0.44 &             0.44 \\
%           &             \textbf{0.30} &      0.28 &             0.26 \\
% \cline{1-4}

% \multirow{2}{*}{diabetes} &             0.40 &  \textbf{0.47} &             0.44 \\
%           &             0.31 &                   0.33 &             0.34 \\
% \cline{1-4}

\multirow{2}{*}{eye}  &             0.60 & \textbf{0.67} &             0.62 \\
          &             0.57 &                   \textbf{0.64} &             0.57 \\
\cline{1-4}

\multirow{2}{*}{heloc} &             0.40 &     \textbf{0.45} &             0.42 \\
          &             0.25 &                   0.25 &             \textbf{0.26} \\
\cline{1-4}

\multirow{2}{*}{magic} &             0.75 &     \textbf{0.80} &    0.73 \\
          &             0.42 &      \textbf{0.55} & 0.46 \\
\cline{1-4}

% \multirow{2}{*}{mushroom} & 0.81 &      \textbf{0.95} &             0.88 \\
          % &             0.74 &  \textbf{0.94} & 0.89 \\
% \cline{1-4}

\multirow{2}{*}{redwine} &  0.52 &  \textbf{0.56} &             0.48 \\
          &             0.69 &      \textbf{0.73} &             0.68 \\
\cline{1-4}

\multirow{2}{*}{tictactoe} &    0.76 &      \textbf{0.79} &             0.78 \\
          &             0.16 &  \textbf{0.19} & 0.18 \\
\cline{1-4}

\multirow{2}{*}{waveform} &     0.49 &  \textbf{0.54} &             0.49 \\
          &             0.10 &      0.14 &     \textbf{0.17} \\
% \cline{1-4}

% \multirow{2}{*}{whitewine} &    0.63 &      \textbf{0.67} &  0.64 \\
%           &             0.55 &      \textbf{0.59} &             0.59 \\
\bottomrule
\end{tabular}

% \caption{Improvement of precision on refinement ($\text{IMD}_\text{6+1}$).}
\label{tab:refinement}
\end{table}
To investigate the effect of the refinement step of IMD (described at the end of Section~\ref{sec:algo:refine}), we compare diff rulesets obtained from three variations of the algorithm --- IMD with maximum depth of 6 ($\text{IMD}_6$), same as in previous experiments; $\text{IMD}_6$ with 1 iteration of refinement ($\text{IMD}_{6+1}$); and IMD with maximum depth of 7 ($\text{IMD}_7$).

Looking at Table~\ref{tab:refinement} (all benchmarks not shown for lack of space), we observe improvement in precision from $\text{IMD}_6$ to $\text{IMD}_{6+1}$ ($11.27$\% on average), and interestingly, also from $\text{IMD}_7$ to $\text{IMD}_{6+1}$ ($4.22$\% on average). The $p$-values from Wilcoxon's test are on the order of $10^{-3}$ for both comparisons, validating the significance of the improvement. The average numbers of rules for the three approaches are 20.93, 28.77, and 41.01 respectively, confirming that $\text{IMD}_{6+1}$ only refines selectively compared to $\text{IMD}_{7}$.

The results demonstrate that selective splitting of impure leaf nodes only in predicted diff regions ($\text{IMD}_{6+1}$), %where the surrogates predict different labels, 
improves precision %(of diff rules) 
compared to regular tree splitting of \textit{all} impure nodes ($\text{IMD}_{7}$). However, this improvement is to be taken with some caution as it comes at the cost of a consistent drop in recall ($15.37$\% from $\text{IMD}_6$ and $25.14$\% from $\text{IMD}_7$ averaged across all benchmarks). Thus we recommend refinement specifically for scenarios requiring high precision difference modelling.

\paragraph{Experimental Conclusions} IMD has close to the same F1-scores as the top methods in our comparison, separate surrogates and the (uninterpretable) Direct GB. At the same time, IMD yields much more concise results, with an order of magnitude fewer diff rules than separate surrogates. This affirms the benefit of sharing nodes in JST, which localizes differences before divergence.
% \textcolor{blue}{
We also see (in SM) how the features deemed important by JST are close to what the models also use in their decision logic via feature importance computations. This establishes our claim that JSTs are able to achieve two things at once: interpretable surrogates that can be compared easily for the two models.
% }
% \textcolor{red}{IMD has better F1-scores than Direct DT and BRCG Diff.~while having a comparable number of diff rules and a more unified JST representation.} 
Refinement further improves the precision of IMD, but at the cost of recall and interpretability. Additional experiments (in SM) %that study the effect of depth on the metrics among the baselines 
also support these conclusions.

\subsection{Case Study}
\label{sec:expt:casestudy}

We conclude by demonstrating a practical application of the method in the fairness area in the advertising domain. Bias in ad campaigns leads to poor outcomes for companies not reaching the right audience, and for customers who are incorrectly targeted. Bias mitigation aims to correct this by changing models to have more equitable outcomes.

Our IMD method can be used to assess the impact of bias mitigation on a model. In this case study, a bias mitigation method was applied to the %privileged 
group of \emph{non}-homeowners who had higher predicted rates of conversion (relative to ground truth). The root node of the JST captures this group. Figure \ref{fig:ad:subtree} shows a part of the JST (full tree in the Appendix). Although the non-homeowner group is already over-predicted, the JST shows that for certain cohorts within the %privileged 
group (those outside the ages of 25-34), conversions are predicted where the model before mitigation would not have. Interpretable model differencing here captures unintended consequences of model alterations.

\begin{figure}[ht]
    \centering
    \includegraphics[width=0.5\columnwidth]{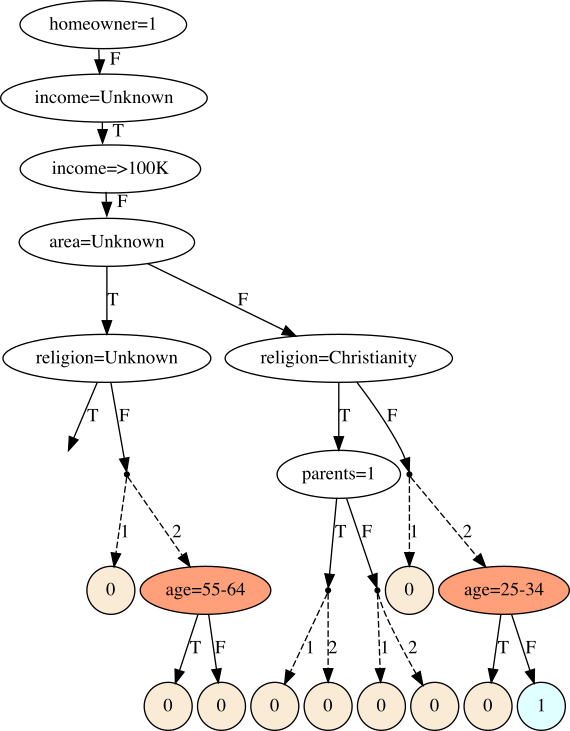}
    \caption{A subtree of the JST showing an unintended increase in predicted conversions after bias mitigation for a cohort of the already over-predicted group of non-homeowners.}
    \label{fig:ad:subtree}
\end{figure}

%%%%%%%%%%%%%%%%%%%%%%%%%%%%%%%%%%%%%%%%%%%
%%%%%%%%%%%%%%%%%%%%%%%%%%%%%%%%%%%%%%%%%%%

\section{CONCLUSION}
\label{sec:concl}

We addressed the problem of interpretable model differencing, localizing and representing differences between ML models for the same task. We proposed JST to provide a unified view of the similarities and dissimilarities between the models as well as a succinct ruleset representation.
Experimental results indicate that the proposed IMD approach yields a favorable trade-off between accuracy and interpretability in predicting model differences.%, close to the accuracy of much more complex diff models while significantly more accurate than interpretable baselines.

The current work is limited to comparing classifiers in terms of $0$-$1$ dissimilarity. Since IMD is based on decision trees, its interpretability is limited to domains where the features are interpretable. While we have chosen to extend greedy decision tree algorithms due to ease and scalability, the resulting JSTs accordingly have no guarantees of optimality.

Future work could seek to address the above limitations. To extend the framework to regression tasks, a potential avenue is to threshold the difference function $D(M_1(x), M_2(x))$ and apply the classification framework presented herein. The problem of interpretable model differencing for images and language remains open. The constituent features for these modalities are generally not interpretable making the diff rulesets uninterpretable without additional considerations.  
  
%%%%%%%%%%%%%%%%%%%%%%%%%%%%%%%%%%%%%%%%%%%
%%%%%%%%%%%%%%%%%%%%%%%%%%%%%%%%%%%%%%%%%%%

\section{Acknowledgements}
This work was partially funded by the European Union’s Horizon Europe research and innovation programme under grant agreement no. 101070568.

%%%%%%%%%%%%%%%%%%%%%%%%%%%%%%%%%%%%%%%%%%%
%%%%%%%%%%%%%%%%%%%%%%%%%%%%%%%%%%%%%%%%%%%

% References
% \clearpage % start references from a new page
\balance % balance the 2 column references last page
\bibliography{haldar_679}

% ----- supplementary -----
\appendix
\onecolumn
\title{Interpretable Differencing of Machine Learning Models\\(Supplementary Material)}
\maketitle

%%%%%%%%%%%%%%%%%%%%%%%%%%%%%%%%%%%%%%%%%%
%%%%%%%%%%%%%%%%%%%%%%%%%%%%%%%%%%%%%%%%%%

\begin{abstract}
    This document contains the following:
    \begin{itemize}
        \item \ref{appdx: bench}: datasets descriptions
        \item \ref{appdx: models}: details of trained models, and their test accuracy values
        \item \ref{appdx: jsts}: an additional example figure of JST
        \item \ref{appdx: repro}: reproducibility checklist
        \item \ref{appdx: pseudocode}: pseudocodes of the algorithms
        
        \item \ref{appdx: expts}: additional experimental results and plots
        \begin{itemize}
            \item \ref{appdx: fid} fidelity comparison of jointly and separately trained surrogates
            \item \ref{appdx: depth} effect of the hyper-parameter \texttt{max\_depth} on F1 and \# rules
            \item \ref{appdx: f1-r-tradeoff} trade-off curves between F1 score and \# rules for interpretable baselines (Direct DT, Separate, and IMD)
            \item \ref{appdx: direct-dt-vs-imd} qualitative comparison between Direct DT and IMD
            \item \ref{appdx: comparison-more} statistical comparison of all baselines on 10 pairs of models per dataset
            \item \ref{appdx: refinement} additional results for refinement on varying depth
            \item \ref{appdx:alpha-expt} effect of the parameter $\alpha$ on metrics
            \item \ref{appdx:perturbation-expt} experiment to demonstrate recovery rate in case of known perturbations
            
        \end{itemize}
        
        \item \ref{appdx:casestudy}: expanded note on the case study
        \item \ref{appdx: full-tables}: full versions of tables in main paper with standard deviations and additional tables
        
    \end{itemize}
\end{abstract}

%%%%%%%%%%%%%%%%%%%%%%%%%%%%%%%%%%%%%%%%%%
%%%%%%%%%%%%%%%%%%%%%%%%%%%%%%%%%%%%%%%%%%

\section{DATASETS}
\label{appdx: bench}

The datasets used for our experiments are reported in Table~\ref{tab:datasets}. The $\% $~of samples belonging to each class label are also listed to show the imbalance in the original datasets.

The Pima Indians Diabetes dataset is from the KEEL repository~\citep{alcala2011keel}, the FICO HELOC dataset is collected from the FICO community~\citep{FICO:2022:Online}, and the bank marketing and eye movements~\citep{eye_movements_dataset} datasets are from OpenML~\citep{OpenML_as_a_whole}. The other 9 tabular datasets are collected from the UCI Repository~\citep{UCIRepo}.

\begin{table}[h]
\centering

\caption{Description of Datasets}

\begin{tabular}{lrrcl}
\toprule
  \textbf{Dataset} &  \textbf{\# Rows} &  \textbf{\# Cols} &  \textbf{\# Labels} &      \textbf{\% Labels} \\
\midrule
    adult &   41034 &      13 &         2 &                               [74.68, 25.32] \\
    bankm &   10578 &       7 &         2 &                                 [50.0, 50.0] \\
 banknote &    1372 &       4 &         2 &                               [55.54, 44.46] \\
 diabetes &     768 &       8 &         2 &                                 [65.1, 34.9] \\
    magic &   19020 &      10 &         2 &                               [35.16, 64.84] \\
    heloc &    9871 &      23 &         2 &                               [52.03, 47.97] \\
 mushroom &    8124 &      22 &         2 &                                 [48.2, 51.8] \\
tictactoe &     958 &       9 &         2 &                               [34.66, 65.34] \\
       bc &     569 &      30 &         2 &                               [37.26, 62.74] \\
 waveform &    5000 &      40 &         3 &                         [33.84, 33.06, 33.1] \\
      eye &   10936 &      26 &         3 &                        [34.78, 38.97, 26.24] \\
whitewine &    4898 &      11 &         7 & [0.41, 3.33, 29.75, 44.88, 17.97, 3.57, 0.1] \\
  redwine &    1599 &      11 &         6 &       [0.63, 3.31, 42.59, 39.9, 12.45, 1.13] \\
\bottomrule
\end{tabular}

\label{tab:datasets}
\end{table}

\newpage
\section{MODELS}
\label{appdx: models}
The model abbreviations and their expanded instantiations (as coded in Scikit-Learn~\citep{scikit-learn}) are shown in Table~\ref{tab:model-params}. Empty instantiations (e.g., \texttt{GaussianNB()}) correspond to default parameter settings.

The test accuracies of the models are listed in Table~\ref{tab:models}.

\begin{table*}[h]
\centering

\caption{Details of the Models}

\begin{tabular}{lr}
\toprule
\textbf{Abbr.} &                                         \textbf{Parameters} \\
\midrule
   LR & \texttt{LogisticRegression(random\_state=1234)} \\
   KN1 & \texttt{KNeighborsClassifier(n\_neighbors=3)} \\
   KN2 & \texttt{KNeighborsClassifier()} \\
  
  MLP1 & \begin{tabular}{@{}c@{}}\texttt{MLPClassifier(alpha=1e-05,hidden\_layer\_sizes=(15,)} \\ \texttt{,random\_state=1234,solver='lbfgs')}\end{tabular}  \\
   
   MLP2 & \begin{tabular}{@{}c@{}}\texttt{MLPClassifier(hidden\_layer\_sizes=(100,100),} \\ \texttt{random\_state=1234)}\end{tabular}  \\

%   MLP2 & \texttt{MLPClassifier(hidden\_layer\_sizes=(100,100),random\_state=1234)} \\

   DT2 & \texttt{DecisionTreeClassifier(max\_depth=10)} \\
   DT1 & \texttt{DecisionTreeClassifier(max\_depth=5)} \\
   GB & \texttt{GradientBoostingClassifier()} \\
   RF1 & \texttt{RandomForestClassifier()} \\
   RF2 & \texttt{RandomForestClassifier(max\_depth=6,random\_state=1234)} \\
   GNB & \texttt{GaussianNB()} \\
   
\bottomrule
\end{tabular}

\label{tab:model-params}
\end{table*}

\begin{table*}[h]
% \scriptsize
% \small
\centering

\caption{Test Accuracy (\%) of Models}

\begin{tabular}{lrrrrrrrrrrr}
\toprule
\textbf{Datasets} & \textbf{LR} &  \textbf{KN1} &  \textbf{DT1} &    \textbf{MLP1} & \textbf{MLP2} & \textbf{DT2} & \textbf{GB} &     \textbf{RF1} & \textbf{KN2} &   \textbf{RF2} & \textbf{GNB} \\
\midrule
adult     &  81.88 &   82.80 &  84.07 &   74.53 &   84.26 &   84.36 &   85.78 &   82.98 &   83.27 &  83.71 &  79.42 \\
bankm     &  74.07 &   73.09 &  77.06 &   71.01 &   66.86 &   76.94 &   80.34 &   80.06 &   74.45 &  78.89 &  71.11 \\
banknote  &  98.30 &  100.00 &  97.33 &  100.00 &  100.00 &   97.82 &   99.51 &   99.51 &  100.00 &  99.51 &  83.74 \\
diabetes  &  77.06 &   71.86 &  74.03 &   71.00 &   68.40 &   69.70 &   80.09 &   75.32 &   72.29 &  76.19 &  75.32 \\
magic     &  79.11 &   79.79 &  82.61 &   81.97 &   84.21 &   84.31 &   86.93 &   88.08 &   80.56 &  85.30 &  72.27 \\
heloc     &  71.51 &   65.23 &  71.61 &   72.08 &   71.34 &   67.45 &   73.06 &   73.09 &   67.93 &  73.16 &  69.38 \\
mushroom  &  99.92 &  100.00 &  99.88 &  100.00 &  100.00 &  100.00 &  100.00 &  100.00 &  100.00 &  99.71 &  96.39 \\
tictactoe &  97.92 &   89.58 &  88.89 &   96.88 &   97.57 &   94.10 &   96.53 &   97.92 &   94.10 &  90.97 &  69.44 \\
bc        &  91.81 &   92.98 &  93.57 &   91.23 &   92.40 &   92.40 &   92.40 &   92.98 &   93.57 &  93.57 &  88.89 \\
waveform  &  86.27 &   80.67 &  75.60 &   83.87 &   83.93 &   75.73 &   85.20 &   84.33 &   80.93 &  83.87 &  79.53 \\
eye       &  44.96 &   47.58 &  51.75 &   45.23 &   44.65 &   56.57 &   61.60 &   66.63 &   47.88 &  57.82 &  43.40 \\
whitewine &  47.28 &   49.86 &  53.54 &   46.94 &   48.16 &   54.63 &   61.22 &   67.14 &   47.41 &  56.87 &  44.29 \\
redwine   &  62.71 &   51.88 &  61.04 &   61.46 &   61.88 &   60.62 &   64.17 &   68.75 &   51.04 &  66.25 &  52.08 \\
\bottomrule
\end{tabular}

\label{tab:models}
\end{table*}

%%%%%%%%%%%%%%%%%%%%%%%%%%%%%%%%%%%%%%%%%%
%%%%%%%%%%%%%%%%%%%%%%%%%%%%%%%%%%%%%%%%%%

\clearpage
\section{EXAMPLE OF JST}
\label{appdx: jsts}

\begin{figure}[h]
    % \centering
    \includegraphics[width=1.01\textwidth,keepaspectratio]{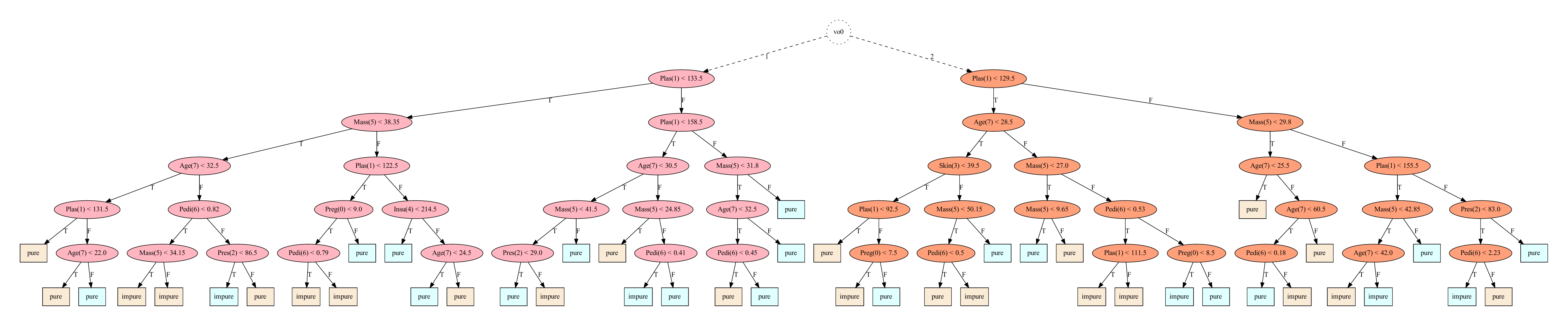}
    \caption{Separate Surrogates (shown side-by-side).}
    \label{fig:separate-ex}
\end{figure}

\begin{figure}[h]
    % \centering
    \includegraphics[width=1.01\textwidth]{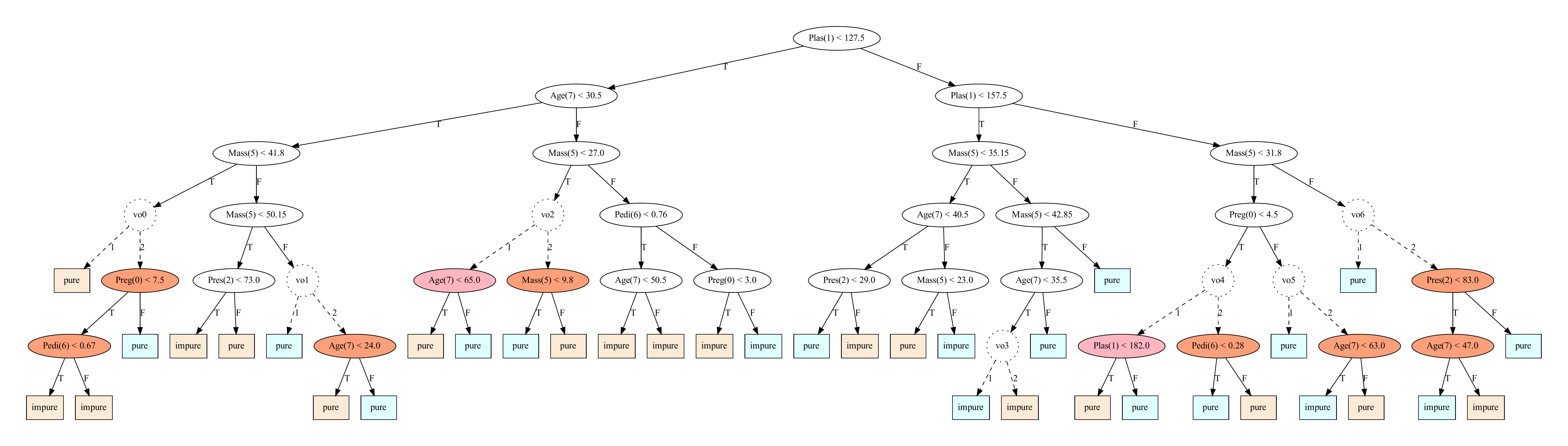}
    \caption{Joint Surrogate Tree (JST)}
    \label{fig:joint-ex}
\end{figure}

Here we give another example of a JST (Figure~\ref{fig:joint-ex}), and also show the separately trained surrogates (Figure~\ref{fig:separate-ex}). For this example, we have picked \textit{diabetes} dataset; and LR, RF1 as the model pair.

In Figure~\ref{fig:separate-ex}, the two individual surrogates are shown side-by-side (one in pink, other in orange), and a single diverging or-node is shown at the top, highlighting that they do not share any split node between them.

Figure~\ref{fig:joint-ex} shows the Joint Surrogate Tree for the same dataset and the model pairs. As noted earlier, the JST first tries to share nodes as much as possible before diverging to two individual subtrees corresponding to the two surrogates at the or-nodes (7 here). In this process the JST also localizes the differences into those 7 or-nodes (9 diff-rules).

As an instance, in Figure~\ref{fig:separate-ex}, we note that the root nodes are $\mathrm{Plas} < 133.5$ and $\mathrm{Plas} < 129.5$ for the pink and orange surrogate trees respectively. The JST in Figure~\ref{fig:joint-ex} however, chooses $\mathrm{Plas} < 127.5$ as the common root node (given by equation~\eqref{eq:impJoint}) that aligns the surrogates to share common decisions, and allows easier comparison of the models.

%%%%%%%%%%%%%%%%%%%%%%%%%%%%%%%%%%%%%%%%%%
%%%%%%%%%%%%%%%%%%%%%%%%%%%%%%%%%%%%%%%%%%

\newpage
% \section{Reproducibility Checklist}
\section{REPRODUCIBILITY CHECKLIST}
\label{appdx: repro}

% \paragraph{Code} We are unable to make code available because it is proprietary due to our institutional obligations. We do provide detailed pseudocode in Appendix~\ref{appdx: pseudocode}.

\paragraph{Code} The code is available at \texttt{https://github.com/Trusted-AI/AIX360}. We also provide detailed pseudocode in Appendix~\ref{appdx: pseudocode}.

\paragraph{Computing infrastructure}
% CPU/GPU specs, amount of memory, OS, relevant software libraries and versions
All the algorithms are implemented in $\texttt{Python 3.8}$, and the experiments are performed in a machine running macOS 12.4 with 2.6 GHz 6-Core Intel Core i7, and 32 GB of memory. We have used $\texttt{scikit-learn==1.1.1}$ in our environment.

\paragraph{ML model parameters} These are given in Table~\ref{tab:model-params}. As mentioned in Section~\ref{sec:expt}, we did not do performance tuning of ML model parameters as our focus is on accurate model differencing and not obtaining the best performing ML models.

%%%%%%%%%%%%%%%%%%%%%%%%%%%%%%%%%%%%%%%%%%
%%%%%%%%%%%%%%%%%%%%%%%%%%%%%%%%%%%%%%%%%%

% \newpage
\section{ALGORITHM PSEUDOCODES}
\label{appdx: pseudocode}

In Algorithm~\ref{algo:dt}, we provide the pseudocode of the plain decision tree fitting algorithm that uses recursion. In Algorithm~\ref{algo:js}, we provide a simplified version of the JST learning algorithm that does not use or-node terminology. In Algorithm~\ref{algo:diff-rules}, the extraction procedure of diff rules from a JST is detailed, following the notations from Section~\ref{section:joint-surrogate-tree}.

\begin{algorithm}[h]
% \scriptsize
\SetAlgoLined
\SetAlgoLined\DontPrintSemicolon
\SetKwProg{Fn}{Function}{}{end}

\KwIn{Samples $X$, and labels $Y$, and also the current $depth$ in the tree.}
\KwOut{A decision tree.}

\Fn{dtfit ($X$, $Y$, $depth=0$)}{
  \If{empty($X$)}{
      \Return none
  }
  \If{all\_same($Y$)}{
	 \Return $Y[0]$
  }
  \If{$depth \ge max\_depth$}{
	\Return majority($Y$)
  }
  $col$, $cutoff$, $ent$ = best\_split($X$, $Y$)\\
  \Return split($col$, $cutoff$, $X$, $Y$, $depth+1$)
}

\Fn{split($col$, $cutoff$, $X$, $Y$, $depth$)}{
  $node.cond = (col < cutoff)$\\
  $sl = X[:,col] < cutoff$\\
  $sr = X[:,col] \ge cutoff$\\
  $node.left = dtfit(X[sl], Y[sl], depth)$\\
  $node.right = dtfit(X[sr], Y[sr], depth)$\\
  \Return $node$  
}

\Fn{best\_split($X$,$Y$)}{
   \Return  	$argmin,\,min_{col, val \in X[:,\,col]}$ entropy($Y$, $col$, $val$)
}

\caption{Decision Tree Algorithm}
\label{algo:dt}
\end{algorithm}

\begin{algorithm}[ph]
% \scriptsize
% \small
\SetAlgoLined
\SetAlgoLined\DontPrintSemicolon
\SetKwProg{Fn}{Function}{}{end}

\KwIn{Samples $X$, labels from $M_1$: $Y1$, labels from $M_2$: $Y2$, and also the current $depth$ in the tree.}
\KwOut{Two decision tree surrogates corresponding to $M_1$ and $M_2$.}

\Fn{jointsurrogate($X$, $Y1$, $Y2$, $depth=0$)}{
  \If{$empty(X) \lor \exists i, all\_same(Yi) \lor depth \ge max\_depth$}{
  
  $\texttt{// base conditions: if one of Y1 or Y2 is pure, diverge to give two separate trees}$\\
  
	\Return $dtfit(X1,Y1,depth)$, $dtfit(X2,Y2,depth)$   
  }
  $col1, cutoff1, ent1$ = $best\_split(X, Y1)$\\
  $col2, cutoff2, ent2$ = $best\_split(X, Y2)$\\
  $col, cutoff, jointent$ = $joint\_best\_split(X, Y1, Y2)$\\

  \uIf{$not\,diverge(ent1,ent2, jointent)$}{
  
  $\texttt{// shared split condition for both}$
  
    $node1.cond = node2.cond = col < cutoff$\\
    $sl = X[:,col] < cutoff$\\
  	$sr = X[:,col] \ge cutoff$\\
	$node1.left, node2.left = jointsurrogate(X[sl], Y1[sl], Y2[sl])$\\
    $node1.right, node2.right = jointsurrogate(X[sr], Y1[sr], Y2[sr])$\\
    \Return $node1, node2$
  }\Else{          	
    $\texttt{// diverge to two separate subtrees (or-node)}$\\
    $\texttt{// uses function from Algorithm~\ref{algo:dt}}$\\
  	\Return $split(col1, cutoff1, X, Y1), split(col2, cutoff2, X, Y2)$
  }
}
  
\Fn{joint\_best\_split($X$,$Y1$,$Y2$)}{
   \Return  	$argmin,min_{col, val \in X[:,col]}$ entropy($Y1$, $col$, $val$) $+$ entropy($Y2$, $col$, $val$)
}
            
\Fn{diverge($ent1$,$ent2$, $jointent$)}{
    $\texttt{// showing simplified divergence criterion}$\\
   \Return $ent1 == 0 \vee ent2 == 0$
}

\caption{Joint Surrogate Tree Learning Algorithm}
\label{algo:js}
\end{algorithm}

\begin{algorithm}[h]
% \scriptsize
% \small
\SetAlgoLined
\SetAlgoLined\DontPrintSemicolon

\SetKwProg{Fn}{Function}{}{end}

\KwIn{A node $v$ in the constructed JST. Set of constraints $path\_cond$ leading to $v$ from $root$ of JST.}
\KwOut{A list of diff-regions $dr$.}

\Fn{diffreg($v$, $path\_cond$)}
{   
    % $dr \gets$ []\\
    
    \uIf{$v \in V_d$}{
        $\texttt{// at a shared decision node}$\\
        $ld$ = \textit{diffreg}($v_T$, $path\_cond \wedge f(v) < t(v)$)\\ 
        $rd$ = \textit{diffreg}($v_F$, $path\_cond \wedge f(v) \ge t(v)$)\\
        \Return $ld \cup rd$\\
    }\Else{
        $\texttt{// at an or-node}$\\
        $dr \gets$ []\\
        \ForEach{$l_1 \in leaves(v^1)$}{
            \ForEach{$l_2 \in leaves(v^2)$}{
                \If {$label(l_1) \neq label(l_2)$}{
                    $c$ = $path\_cond \wedge pc(l_1) \wedge pc(l_2)$\\
                    \If {$non\_empty(c)$}{
                        $dr.add(c)$
                    }  
                } 
            }
        }
        \Return $dr$\\
    }
}
\textit{diffreg}($JST.root$, $True$)
\caption{Diff-Regions from JST}
\label{algo:diff-rules}
\end{algorithm}

%%%%%%%%%%%%%%%%%%%%%%%%%%%%%%%%%%%%%%%%%%
%%%%%%%%%%%%%%%%%%%%%%%%%%%%%%%%%%%%%%%%%%

% Additional Experiments

\clearpage
% \section{Additional Experimental Results}
\section{ADDITIONAL EXPERIMENTAL RESULTS}
\label{appdx: expts}

\subsection{Fidelity Comparison}
\label{appdx: fid}
We also investigate if jointly training surrogates affects the fidelity (fraction of samples for which the surrogate predictions match with the original model's prediction for a set of instances) of surrogates to the original models. We compute fidelity of the surrogates, $\hat{M}_1$ and $\hat{M}_2$ (corresponding to $M_1$ and $M_2$), on the held-out $\Dtest$.
It can be seen in Table~\ref{tab:fidelity-comparison} that the corresponding fidelity values are very close to each other.

\paragraph{Conclusion} This indicates that the proposed method of jointly approximating two similar models via JSTs, achieves a way to incorporate knowledge from both models (by preferring shared nodes) without harming the individual surrogates' faithfulness to their respective models.

\begin{table*}[h]
\centering

\caption{Fidelity (\%) values for surrogate $\hat{M}_i$ are comparable for separate and joint training procedures.}

\begin{tabular}{llrrrr}
\toprule
          &             & \multicolumn{2}{c}{\textbf{Separate}} & \multicolumn{2}{c}{\textbf{Joint}} \\
% \cline{3-6}
\cmidrule(r){3-4} \cmidrule(l){5-6}

\textbf{Dataset} & \textbf{$M_1$ vs. $M_2$} &  $\hat{M}_1$ &  $\hat{M}_2$ &            $\hat{M}_1$ &  $\hat{M}_2$ \\

% \textbf{D} & \textbf{M1 vs. M2} &              &       &                 &       \\
\midrule
\multirow{2}{*}{adult} & max MLP1-GB &         99.996 &  96.920 &           99.976 &  96.920 \\
          & min MLP2-DT2 &         91.956 &  98.300 &           91.896 &  98.162 \\
\cline{1-6}
\multirow{2}{*}{bankm} & max MLP2-GB &         88.766 &  92.174 &           88.670 &  91.708 \\
          & min MLP1-GNB &         90.328 &  95.432 &           89.140 &  93.868 \\
\cline{1-6}
\multirow{2}{*}{banknote} & max KN1-GNB &         98.056 &  98.204 &           98.396 &  97.620 \\
          & min LR-DT1 &         98.156 &  97.864 &           97.526 &  97.864 \\
\cline{1-6}
\multirow{2}{*}{bc} & max DT1-GNB &         94.270 &  96.610 &           93.216 &  96.142 \\
          & min KN2-RF2 &         95.322 &  93.450 &           92.982 &  93.686 \\
\cline{1-6}
\multirow{2}{*}{diabetes} & max MLP2-GB &         79.828 &  84.416 &           79.222 &  83.984 \\
          & min RF1-GNB &         80.866 &  86.668 &           77.316 &  87.878 \\
\cline{1-6}
\multirow{2}{*}{eye} & max RF1-GNB &         56.178 &  86.912 &           48.962 &  86.010 \\
          & min LR-MLP1 &         78.982 &  79.038 &           76.282 &  76.974 \\
\cline{1-6}
\multirow{2}{*}{heloc} & max KN1-RF2 &         75.022 &  93.592 &           75.416 &  93.396 \\
          & min GB-RF1 &         92.896 &  79.014 &           92.760 &  79.636 \\
\cline{1-6}
\multirow{2}{*}{magic} & max RF1-GNB &         86.934 &  96.848 &           86.156 &  96.540 \\
          & min MLP2-DT2 &         91.376 &  91.494 &           90.422 &  90.754 \\
\cline{1-6}
\multirow{2}{*}{mushroom} & max KN1-GNB &         99.984 &  98.844 &          100.000 &  98.426 \\
          & min RF2-GNB &         99.952 &  98.844 &           99.952 &  98.198 \\
\cline{1-6}
\multirow{2}{*}{redwine} & max RF1-KN2 &         63.042 &  62.292 &           63.918 &  62.626 \\
          & min KN1-GNB &         55.998 &  79.582 &           53.498 &  76.918 \\
\cline{1-6}
\multirow{2}{*}{tictactoe} & max LR-GNB &         89.652 &  91.736 &           90.554 &  93.818 \\
          & min DT2-KN2 &         90.970 &  88.544 &           92.152 &  89.098 \\
\cline{1-6}
\multirow{2}{*}{waveform} & max LR-DT1 &         80.334 &  97.774 &           81.600 &  98.252 \\
          & min MLP1-RF2 &         77.986 &  83.918 &           78.746 &  84.480 \\
\cline{1-6}
\multirow{2}{*}{whitewine} & max RF1-GNB &         59.838 &  80.842 &           58.490 &  78.736 \\
          & min LR-KN2 &         90.272 &  55.658 &           89.360 &  53.526 \\
\bottomrule
\end{tabular}

\label{tab:fidelity-comparison}
\end{table*}

\subsection{Effect of Maximum Depth}
\label{appdx: depth}

In the main paper, we have set the maximum depth hyper-parameter to 6 for separate surrogates, IMD and Direct DT. That choice was made to achieve a favourable trade-off between accuracy and interpretability, and also for ease of inspection of the resulting trees as the maximum decision path length we had to look at was limited to 6.

\begin{figure}[ph]
    \centering
    \includegraphics[width=\textwidth]{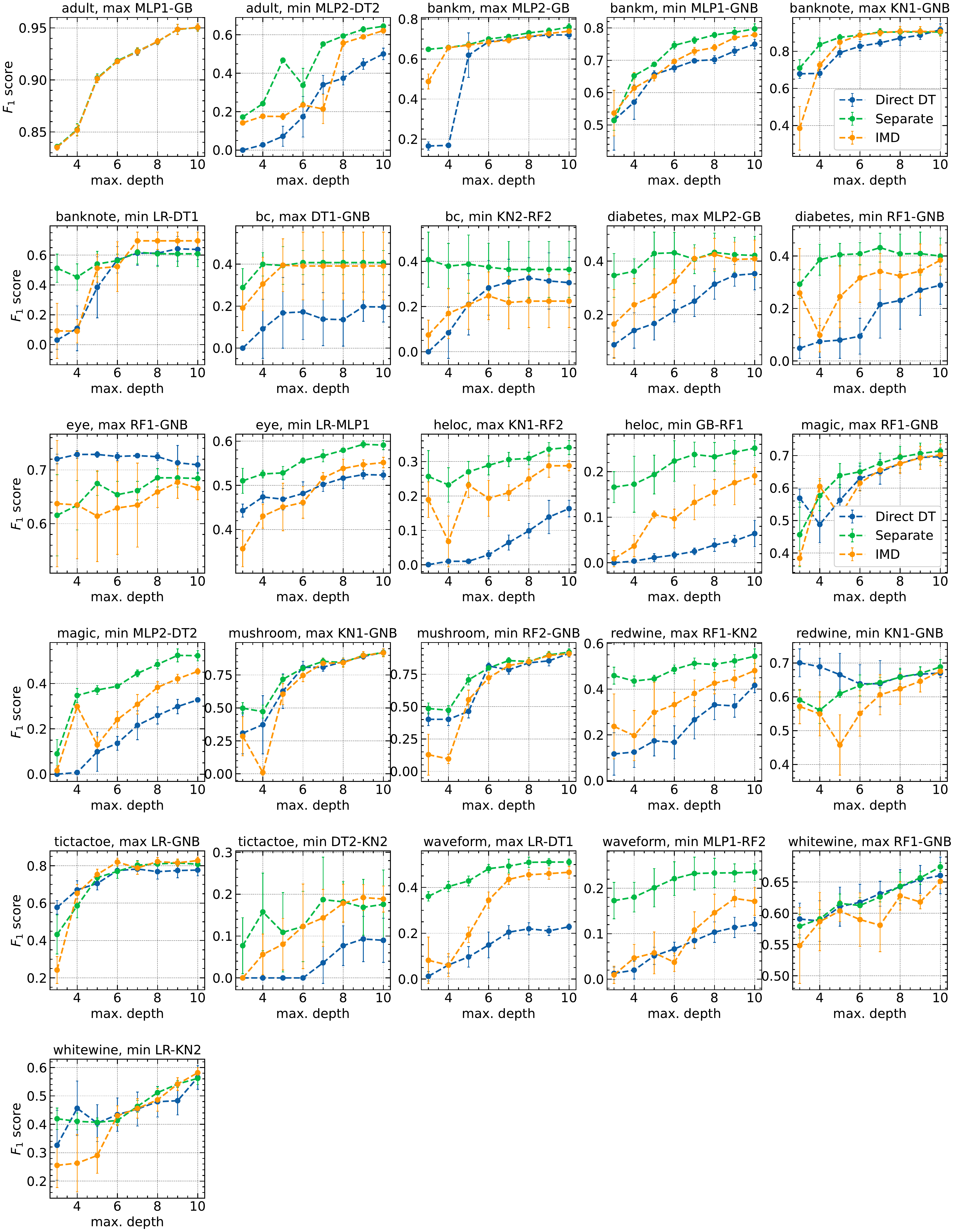}
    \caption{Effect of varying maximum depth on F1-scores for Separate Surrogates (green), IMD (orange) and Direct DT (blue). Each plot in the figure corresponds to a dataset and a model pair, as written on the title. The vertical bars around each point indicate the standard deviation over the 5 runs.
    Typically, we observe the trend separate surrogates on top, IMD in the middle, and Direct DT on the bottom. The values corresponding to max. depth = $6$ are reported in main paper.}
    \label{fig:f1-vs-depth}
\end{figure}

\begin{figure}[ph]
    \centering
    \includegraphics[width=\textwidth]{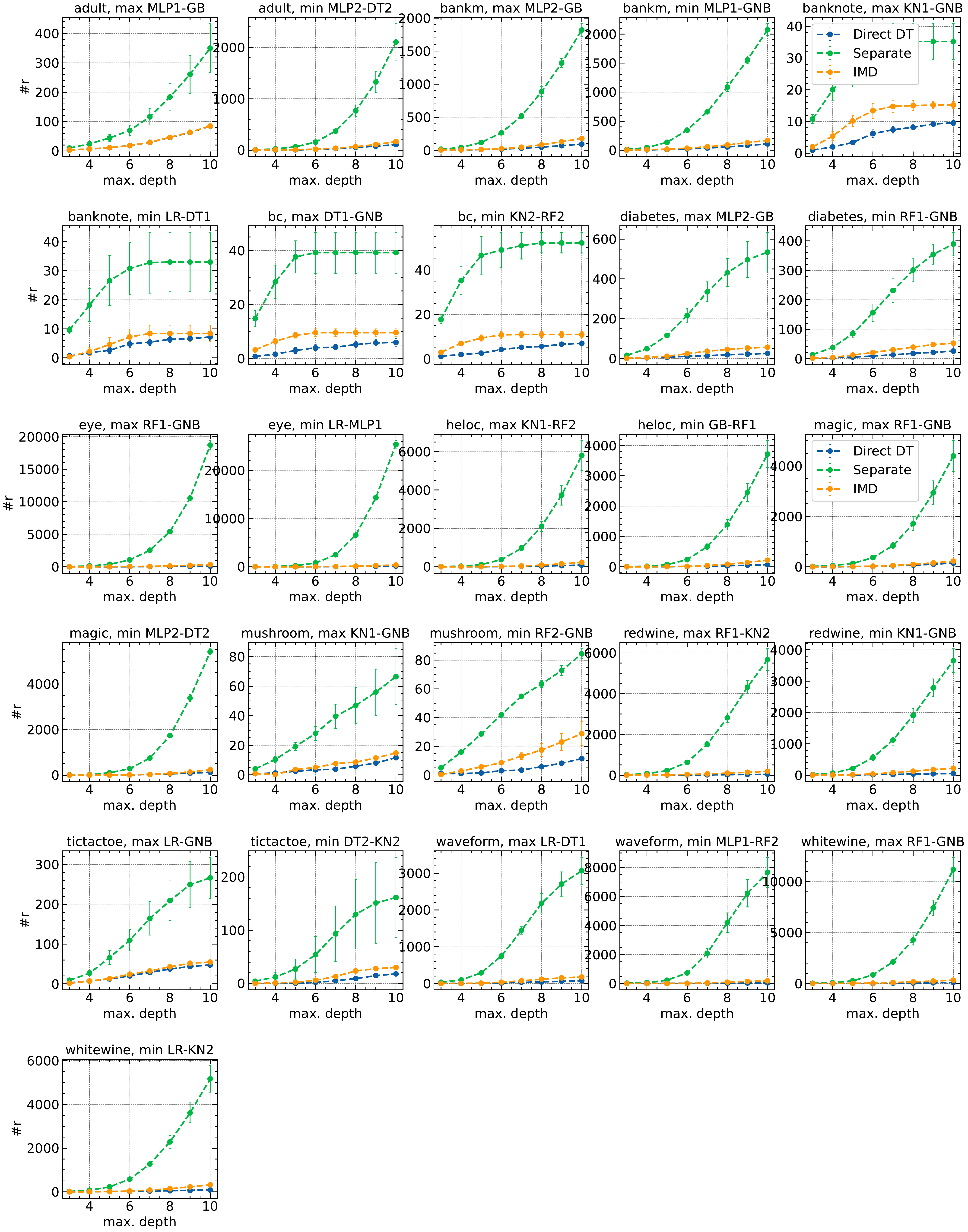}
    \caption{Interpretability (number of rules in the diff ruleset) for Separate Surrogates (green), IMD (orange) and Direct DT (blue) on varying maximum depth. Each plot in the figure corresponds to a dataset and a model pair, as written on the title. The vertical bars around each point indicate the standard deviation over the 5 runs.
    As seen earlier, no. of rules for separate surrogates rises quickly with depth, whereas for IMD and Direct DT it is almost always below 50. The values corresponding to max. depth = $6$ are reported in main paper.}
    \label{fig:rules-vs-depth}
\end{figure}

Now we study how varying the maximum depth of the trees in the methods affects the quality (F1-score) and interpretability (no.~of rules) of the resulting diff ruleset. We have varied the maximum depth from 3 to 10, and plotted the F1-score in Figure~\ref{fig:f1-vs-depth}, and \# rules in Figure~\ref{fig:rules-vs-depth}, and elaborate on some observable trends below.

Looking at Figure~\ref{fig:f1-vs-depth}, we observe the trend that separate surrogates achieves the highest F1 scores, followed by IMD, and then Direct DT, for most (19 out of 26) benchmarks. Typically, we also see rising values of F1 with increasing max.~depth. 

Interestingly, for benchmarks where the fraction of diff samples was high (e.g., \textit{eye} max ($0.56$), \textit{redwine} min ($0.52$), \textit{whitewine} max ($0.53$)), Direct DT mostly outperformed surrogate based approaches, but on increasing depth they did catch up. On the other hand, when the fraction of diff samples were low (e.g., \textit{tictactoe} min ($0.06$), \textit{heloc} max ($0.23$), \textit{bankm} max ($0.26$), \textit{magic} min ($0.11$)) Direct DT gave close to 0 precision and recall initially, whereas both IMD and separate surrogate methods identified regions of differences. 

\paragraph{Conclusion} The above observation also highlights that direct approaches treat difference modelling as an imbalanced classification problem only, and have the same drawbacks. Also see below (Appendix~\ref{appdx: direct-dt-vs-imd}) for a qualitative comparison.

In Figure~\ref{fig:rules-vs-depth}, we observe that for separate surrogates the number of rules values are orders of magnitude higher than both IMD and Direct DT, and rises quickly (or saturates for smaller datasets, eg. \textit{banknote} and \textit{bc}) on increasing the maximum depth. Here, we observe that for most benchmarks, F1 values of IMD at maximum depth of 10 is very close to that of separate surrogates, but the \# rules gap between them is very large, affirming the compactness of representation of JSTs.

\subsection{ACCURACY-INTERPRETABILITY TRADE-OFF FOR INTERPRETABLE BASELINES}
\label{appdx: f1-r-tradeoff}
\begin{figure}[ph]
    \centering
    \includegraphics[width=\textwidth]{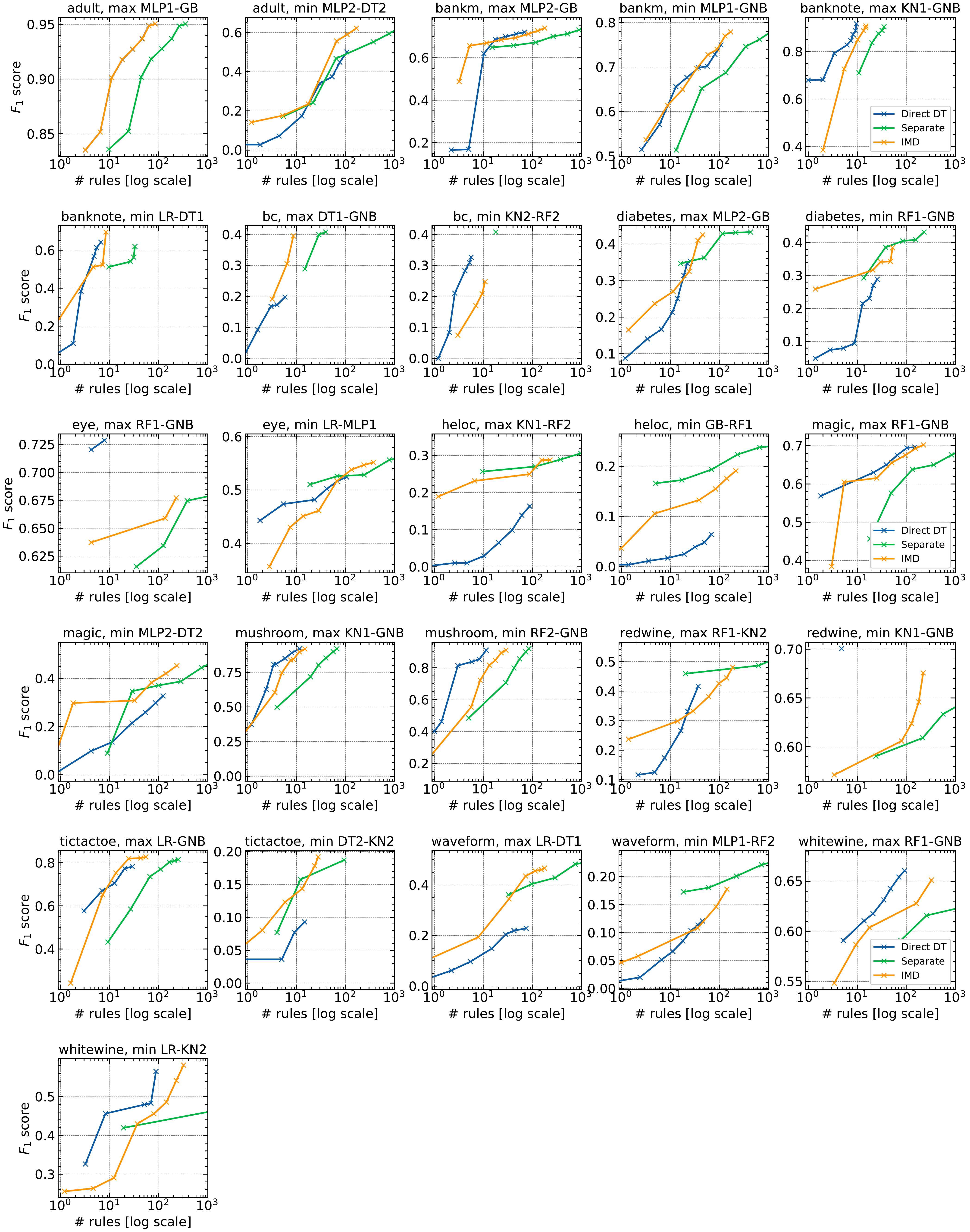}
    \caption{Trade-off between F1-score and \# rules (in log scale, trimmed after $10^3$) for interpretable baselines. Pareto efficient points are connected by line segments and shown in the figure. For some dataset and model pairs (e.g., whitewine, max RF1-GNB), the curve for separate surrogates approach extends beyond the $10^3$ mark.}
    \label{fig:f1-vs-rules}
\end{figure}

Here we compare the trade-off between F1 score and \# rules for interpretable or rule-based difference prediction baselines, namely, Direct DT, separate surrogates, and the proposed method IMD. We show the trade-off plots for different dataset and model pair combinations in Figure~\ref{fig:f1-vs-rules}. These plots are generated by plotting the F1-scores against \# rules, that were obtained on varying the maximum depth hyper-parameter in the last experiment (Appendix~\ref{appdx: depth}). Note that we only connect pareto-efficient points, i.e., those not dominated by points with both higher F1 score and lower \# rules, with line segments for better visualization.
We also plot the rules axis in log scale due to the wide range of \# rules in the separate approach.

Looking at the plots, we see that Direct DT and IMD almost always achieve better trade-off than Separate. On 20 out of the 26 cases shown in Figure~\ref{fig:f1-vs-rules}, the curves for Separate (in green) stay below (or much to the right) of those for Direct DT or IMD (in orange or blue). Also, often the green curves extend beyond $10^3$ rules highlighting its massive complexity.

The trade-offs for Direct DT and IMD are however more competitive, and IMD curve (in orange) is better or is as good as Direct DT for 14 out of the 26 cases. To differentiate betweeen them further, below we have done a qualitative study (Appendix~\ref{appdx: direct-dt-vs-imd}), and also tabulate an extended summary statistics for 10 pairs of models per dataset (Appendix~\ref{appdx: comparison-more}).

\subsection{QUALITATIVE COMPARISON OF DIRECT DT \& IMD}
\label{appdx: direct-dt-vs-imd}

Next we discuss the differences between Direct DT and our proposed method IMD. In the main paper, we noted that the design of JST helps to localize the model differences in the context of overall decision logic of the models (penultimate paragraph in Section~\ref{sec: intro}, also in Figure~\ref{fig:jst}). Here we demonstrate that the JST's internal decision logic (in terms of the features used) is indeed closer to that of the models, while Direct DT's is not.

\begin{figure}[ph]
    \centering
    \includegraphics[width=\textwidth]{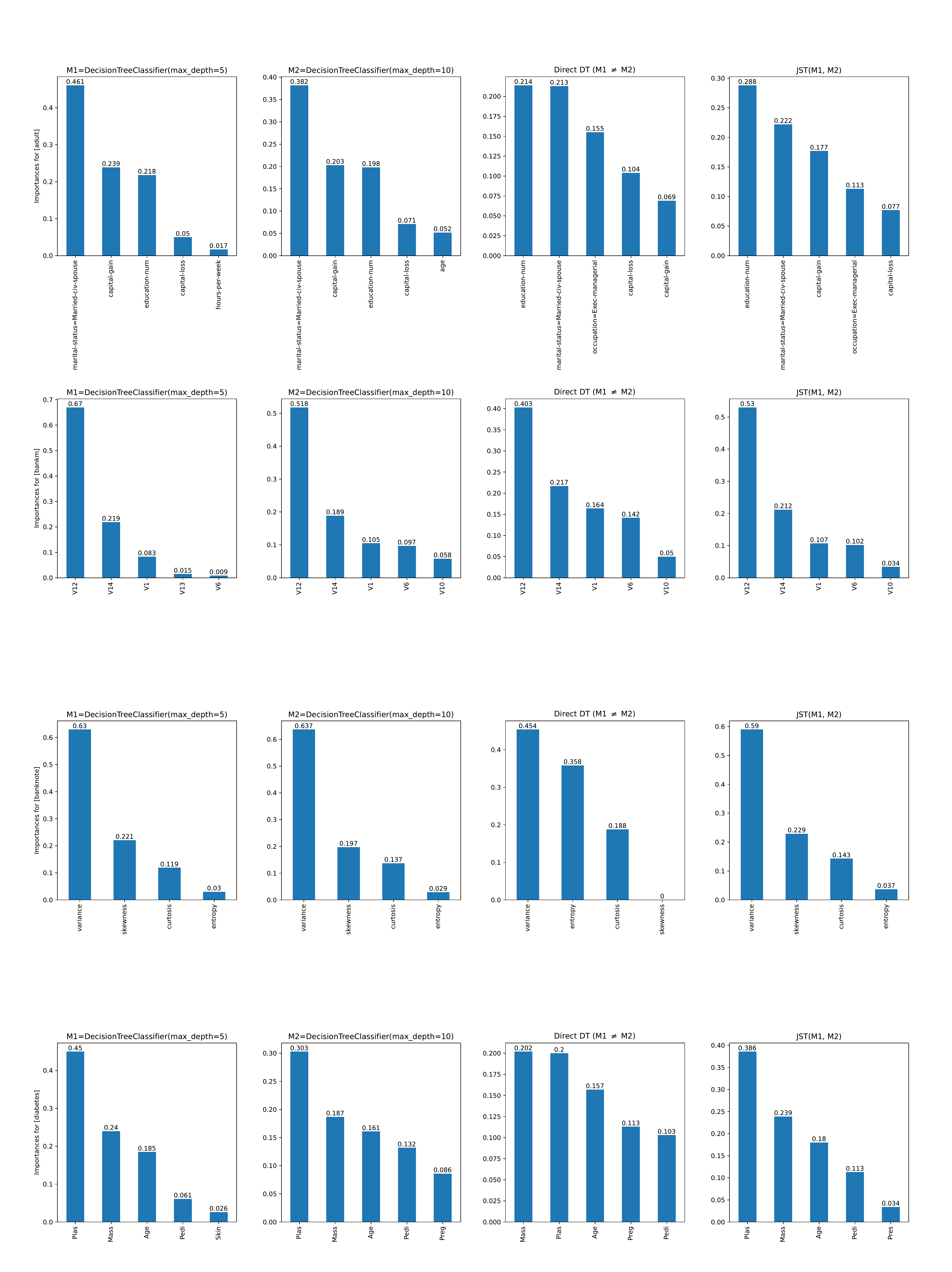}
    \caption{Feature importances for decision tree models (M1 and M2), Direct DT, and IMD (JST) for \textit{adult, bankm, banknote, and diabetes} datasets (each row corresponds to a single dataset.) The importances for JST are much closer to that of the models, demonstrating their decision logics are similar.}
    \label{fig:fi-1}
\end{figure}

\begin{figure}[ph]
    \centering
    \includegraphics[width=\textwidth]{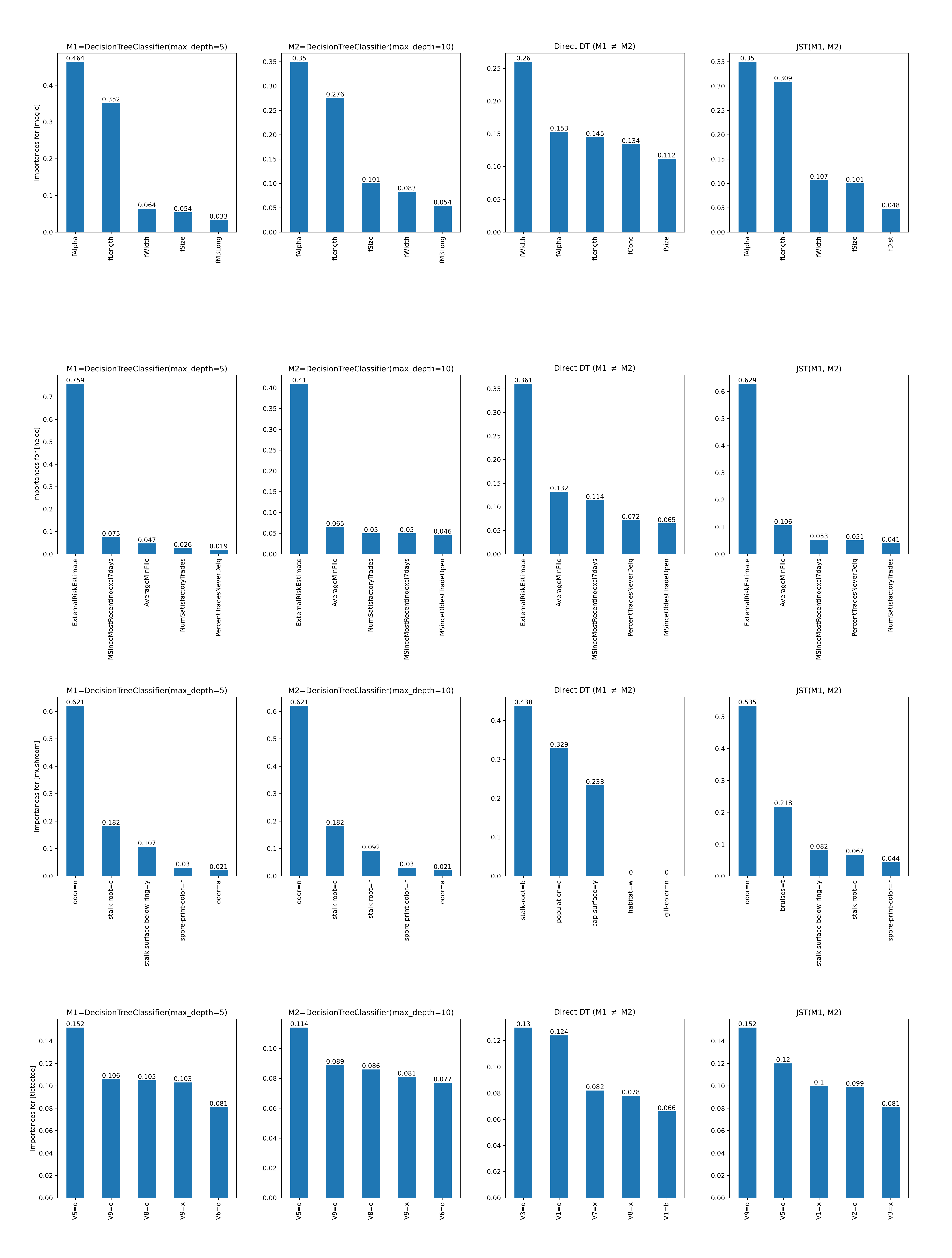}
    \caption{Feature importances for decision tree models (M1 and M2), Direct DT, and IMD (JST) for \textit{magic, heloc, mushroom, and tictactoe} datasets.}
    \label{fig:fi-2}
\end{figure}

\begin{figure}[ph]
    \centering
    \includegraphics[height=\textheight,width=\textwidth]{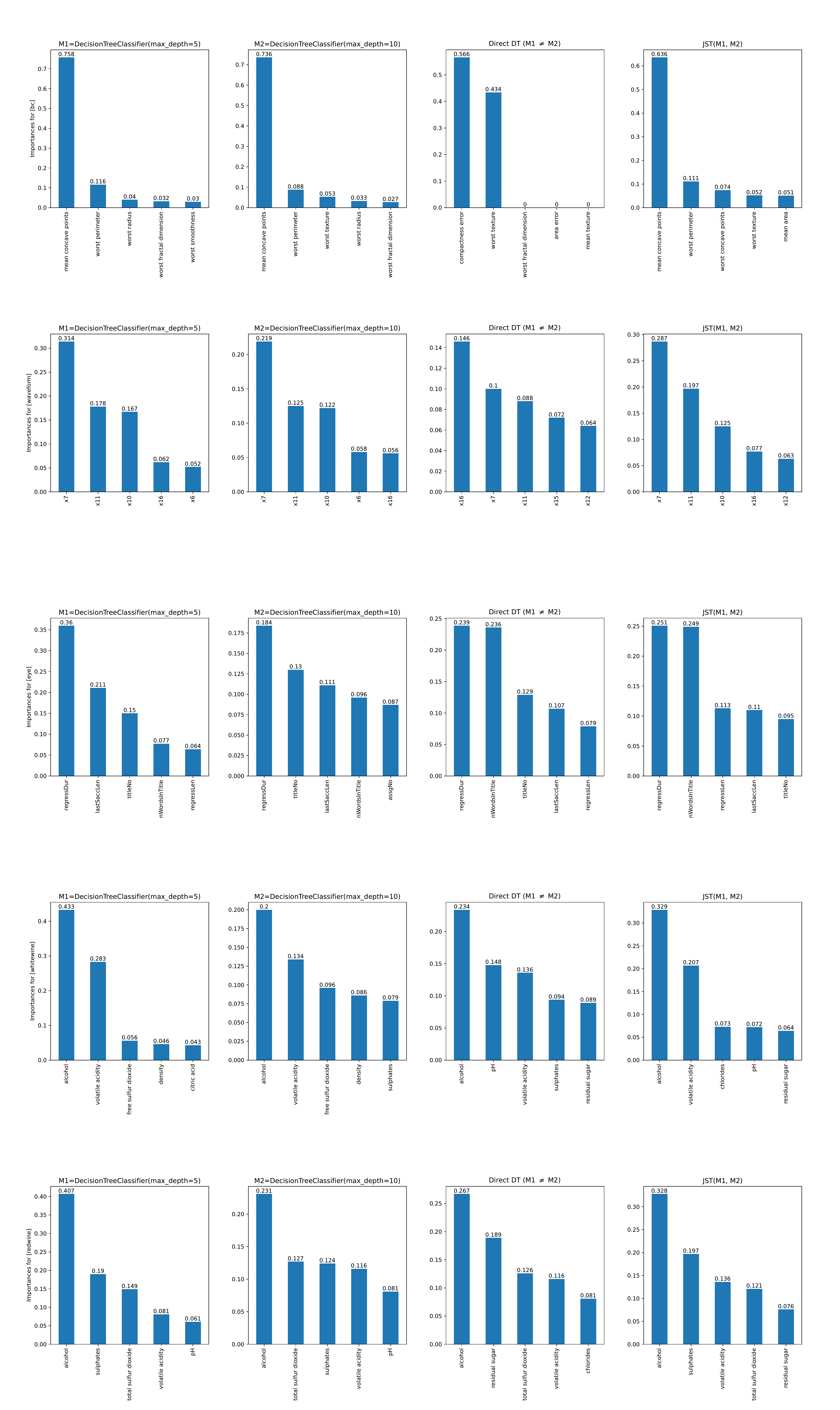}
    \caption{Feature importances for decision tree models (M1 and M2), Direct DT, and IMD (JST) for \textit{bc, waveform, eye, whitewine, and redwine} datasets.}
    \label{fig:fi-3}
\end{figure}

We utilise feature importance scores in decision trees (defined as the total impurity decrease brought in by a feature) to understand the top 5 most important features in Direct DT and JST. We also choose decision trees as the original two models under comparison, i.e., DT1 and DT2, for each dataset. In Figures~\ref{fig:fi-1}, \ref{fig:fi-2}, and \ref{fig:fi-3}, we show for one dataset in each row, the top 5 features used by the models DT1 and DT2 (first two columns), Direct DT differencing (third column), and finally for the JST (the first step for IMD) built on them.

A general observation across all such rows is that the feature importance scores for JST tend to be closer to the models. As an example, for the 4-feature \textit{banknote} dataset, we see both DT1 and DT2 weigh the features \textit{variance, skewness, curtosis}, and \textit{entropy} (in that order), which is same as JST but different from how Direct DT places them: \textit{variance, entropy, curtosis}, and \textit{skewness}. It is also interesting to note that while the feature \textit{skewness} is rated second in the list by the models and JST, Direct DT does not use it at all (indicated by the 0 value).

We also observe that in some cases, the feature with the highest importance value in models is also placed at the top by JST, while it does not appear in the top 5 features for Direct DT. The features \textit{odor=n} for \textit{mushroom} dataset (3rd row in Figure~\ref{fig:fi-2}), and \textit{mean concave points} for \textit{bc} dataset (1st row in Figure~\ref{fig:fi-3}) are two such examples.

\paragraph{Conclusion} The above observations demonstrate that JSTs (and thus IMD) indeed faithfully capture the decision making process of the original models before localizing the differences, while Direct DT focuses solely on identifying the differences without placing them in the context of the models.

\subsection{Comparison of all baselines on More Benchmarks}
\label{appdx: comparison-more}
In Section~\ref{sec:expt:alt}, we compared IMD against separate surrogates, Direct DT, Direct GB and BRCG Diff.~on the basis of the quality of prediction of the true dissimilarity $D$ (F1 scores) for only 2 model pairs per dataset. Here we choose 10 model pairs per dataset (sorted in decreasing order of accuracy gaps), and repeat the same experiment to empirically verify the trend amongst the 5 algorithms for difference modelling. To include BRCG Diff.,~we compare all of them on the 9 binary classification datasets and show the summary in Table~\ref{tab:more-comparison}. In total, we have 90 dataset and model pair combinations.

We show the no.~of times IMD achieved better F1 score than the other algorithms (\# greater), the average \% change in F1 score on going from IMD to the other algorithms, and $p-$values obtained from Wilcoxon's signed rank test. We note a caveat in this analysis that even though the model pairs for different datasets are independent; however, for a given dataset, if the same model appears in multiple model pairs, they are not truly independent. So the $p-$values may be overstated.

We also performed Friedman’s test which confirmed significant difference amongst the algorithms with $p$- value $10^{-23}$. The mean ranks found are $3.23$, $2.16$, $3.78$, $2.1$, $3.73$ for IMD, Separate, Direct DT, Direct GB, and BRCG Diff.~respectively.

\paragraph{Conclusion} This analysis validates that the proposed IMD approach is close to the accuracy of much more complex diff models (Separate and Direct GB), while significantly more accurate than interpretable baselines (Direct DT and BRCG Diff.).

\begin{table}[h]
\centering

\caption{Summary statistics of comparisons on more benchmarks. All observed differences are significant with respect to the Holm-corrected thresholds $0.0125$, $0.017$, $0.025$, $0.05$.}

\begin{tabular}{ccccc}
\toprule
{} & \multicolumn{4}{c}{\textbf{IMD} vs.} \\
{Statistic} &  \textbf{Separate} & \textbf{Direct DT} & \textbf{Direct GB} & \textbf{BRCG Diff.} \\
\midrule
\# IMD has greater F1 &     $15/90$ &     $62/90$ &     $28/90$ &     $52/90$ \\
\% change from IMD   &    $19.67\%$ &   $-14.38\%$ &     $0.43\%$ &   $-13.49\%$ \\
Wilcoxon's $p$-value &  $3.06e-11$ &  $2.60e-06$ &   $0.00028$ &  $5.09e-06$ \\
\bottomrule
\end{tabular}

\label{tab:more-comparison}
\end{table}

\subsection{Further results for Refinement}
\label{appdx: refinement}
In this section, we show the effect of refinement at a general height $h$, with 1 and 2 iterations of selective refinement, for $h=3\,,4\,$ and $5$.

\begin{figure}[ph]
    \centering
    \includegraphics[width=\textwidth]{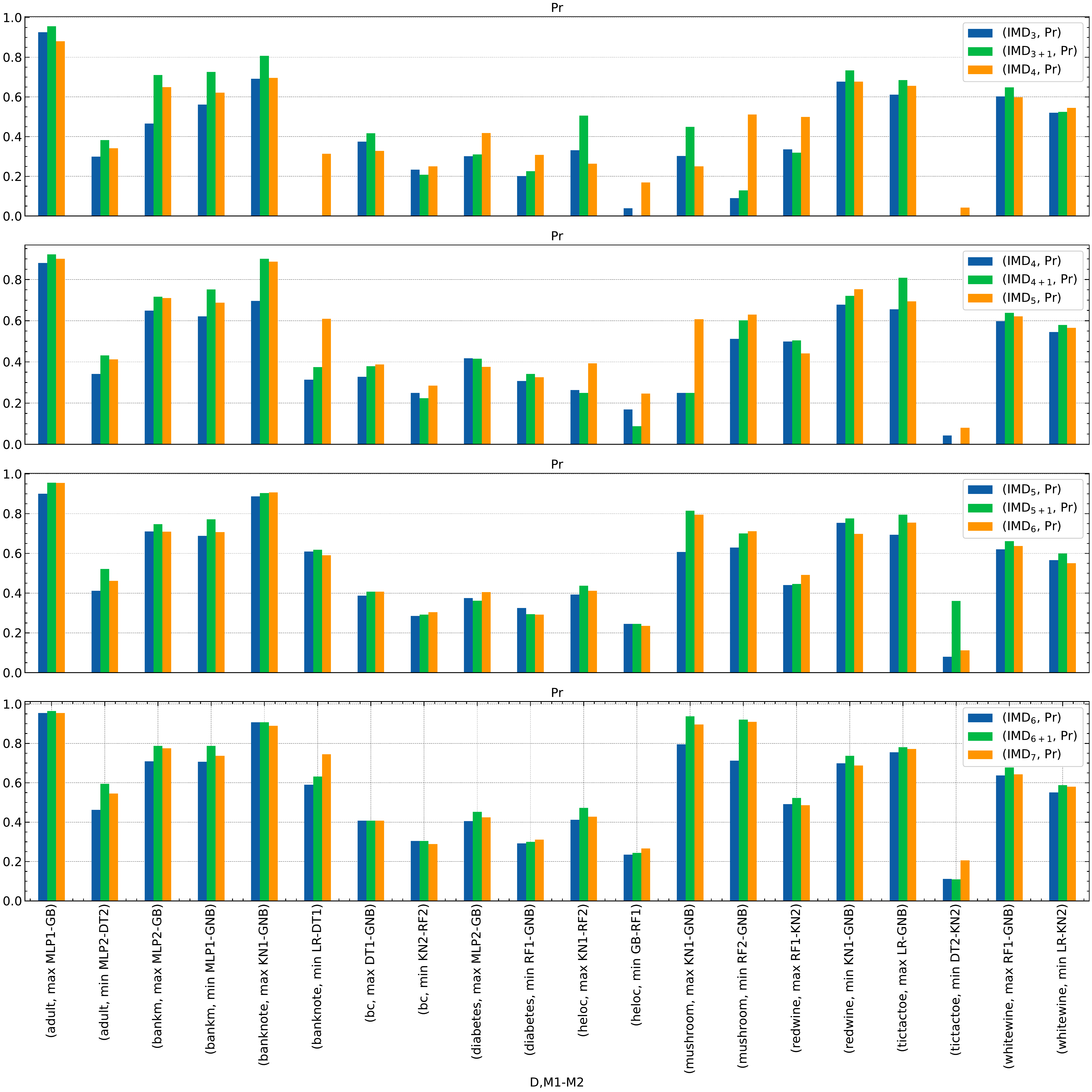}
    \caption{Precision values for $\textrm{IMD}_{h}$, $\textrm{IMD}_{h+1}$ (\textit{1 step refinement}) and $\textrm{IMD}_{h+1}$ (without ref.). Each group of 3 bars (blue, green and orange) in the plot corresponds to the three methods for a dataset and model pair. The trend is, the middle green bar is higher than both its left and right bars, for most benchmarks.}
    \label{fig:ref-h+1}
\end{figure}

In Figure~\ref{fig:ref-h+1}, we compare the precision values of $\textrm{IMD}_{h}$, $\textrm{IMD}_{h+1}$ (with 1 step refinement from height $h$) and $\textrm{IMD}_{h+1}$ (without refinement). As can be seen in the bar plots, the middle green bar corresponding to $\textrm{IMD}_{h+1}$ (with 1 level refinement from height $h$) achieves higher precision than both the IMD versions at height $h$, and $h+1$ without refinement.

\begin{figure}[ph]
    \centering
    \includegraphics[width=\textwidth]{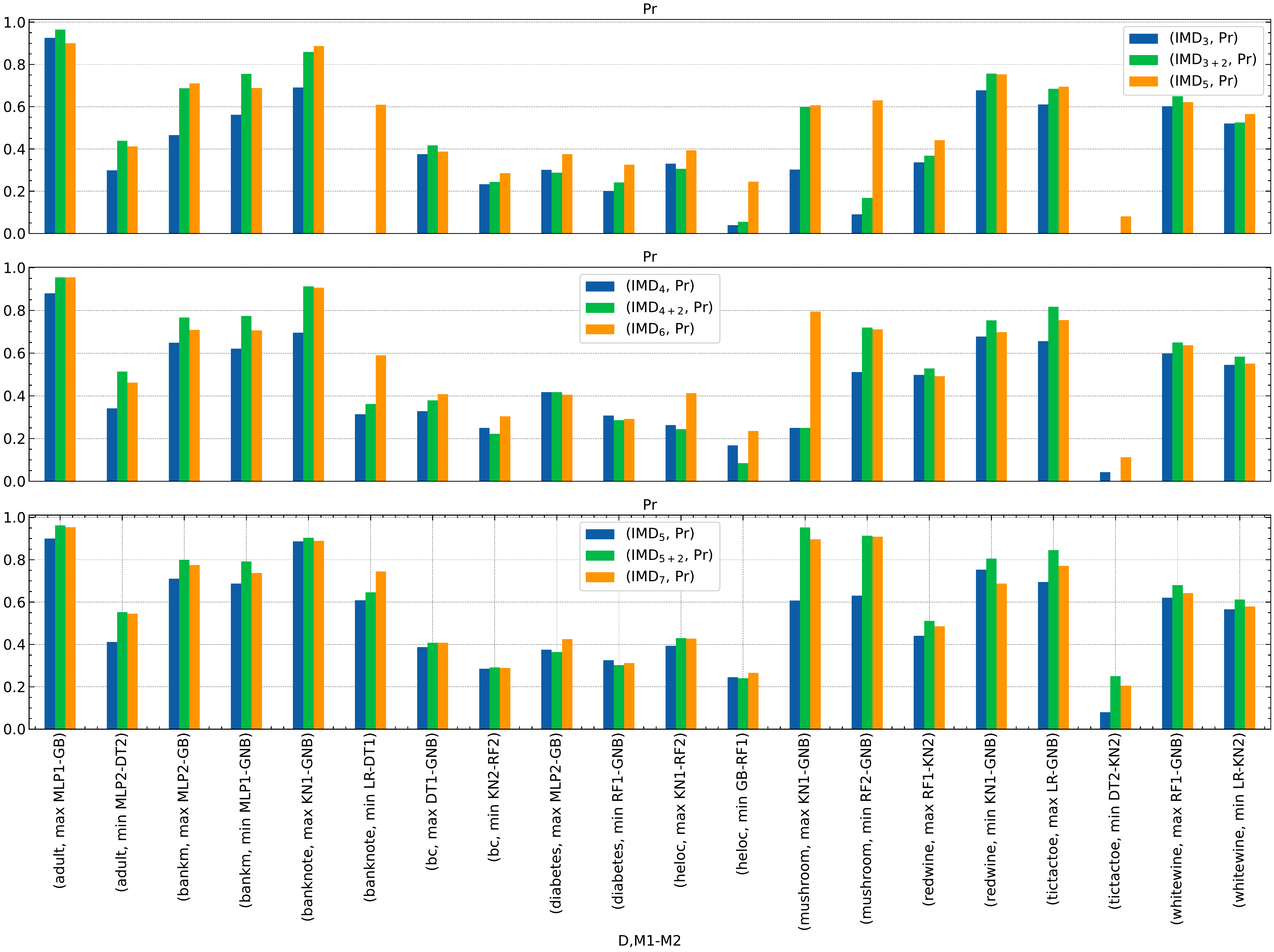}
    \caption{Precision values for $\textrm{IMD}_{h}$, $\textrm{IMD}_{h+2}$ (\textit{2 step refinement}) and $\textrm{IMD}_{h+2}$ (without ref.). Each group of 3 bars (blue, green and orange) in the plot corresponds to the three methods for a dataset and model pair. The trend is, the middle green bar is higher than both its left and right bars, for most benchmarks.}
    \label{fig:ref-h+2}
\end{figure}

We see similar trends for 2 steps of refinement in Figure~\ref{fig:ref-h+2} for $\textrm{IMD}_{h}$, $\textrm{IMD}_{h+2}$ (with 2 step refinement from height $h$) and $\textrm{IMD}_{h+2}$ (without refinement).

\subsection{Effect of the parameter $\alpha$}
\label{appdx:alpha-expt}

In the main paper, we have used the simplified divergence criterion~\eqref{eq:divCond2} instead of~\eqref{eq:divCond1} which was parameterized by $\alpha$. In this experiment, we study the trend of F1 scores and \# rules against $\alpha$ (ranging from $0$ to $1$), with a fixed \texttt{max\_depth} of 6.

We emphasize again that the end point $\alpha=1.0$ corresponds to separate surrogate approach, and $\alpha$ close to 0 would encourage more and more sharing of nodes.

\begin{figure}[ph]
    \centering
    \includegraphics[width=\textwidth]{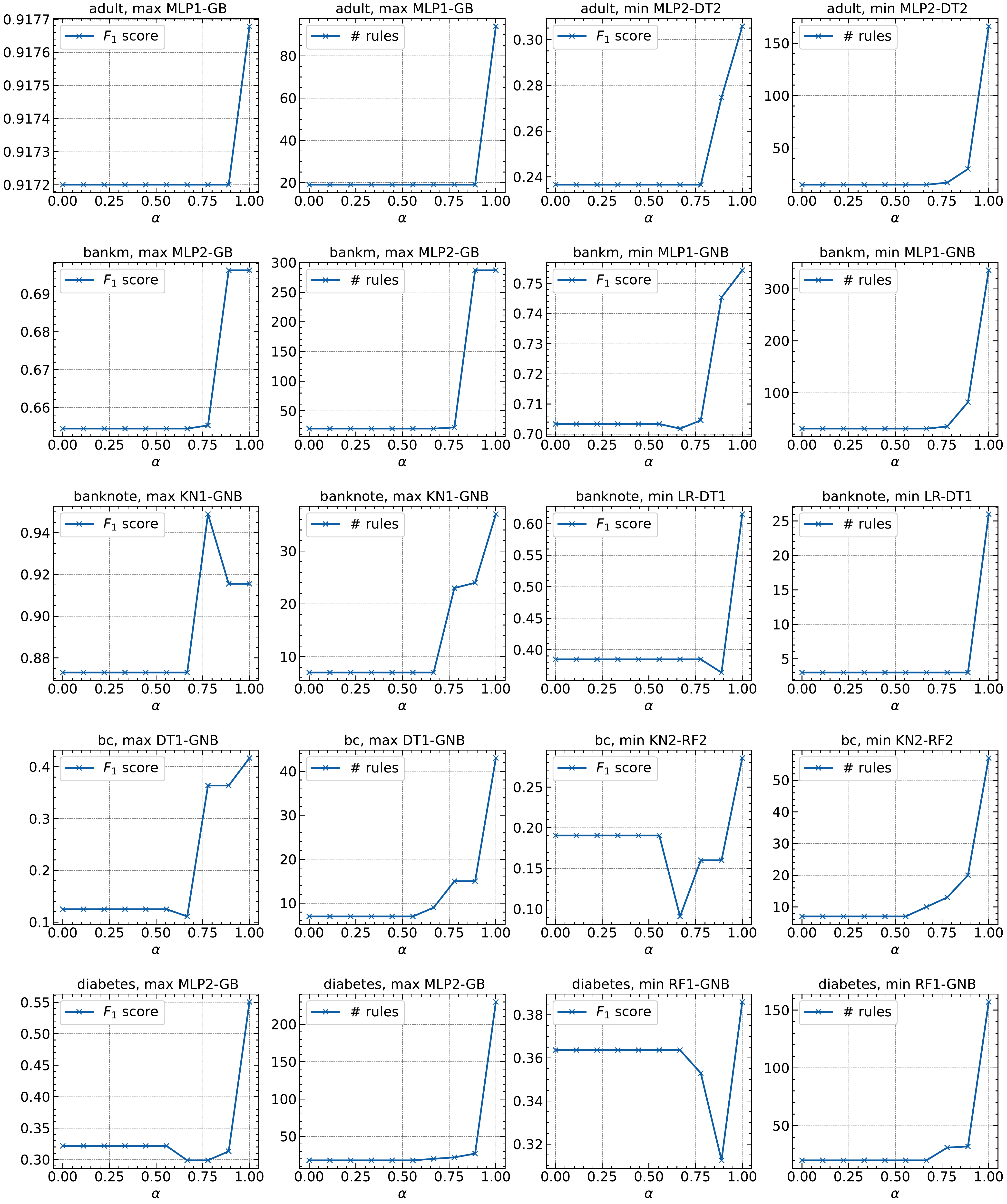}
    \caption{Effect of $\alpha$ on metrics for IMD. The algorithm is sensitive to $\alpha$ close to $1$ and quickly makes a transition from conjoined trees to completely separate trees.}
    \label{fig:metrics-vs-alpha}
\end{figure}

\begin{figure}[ph]
    \centering
    \includegraphics[width=0.9\textwidth]{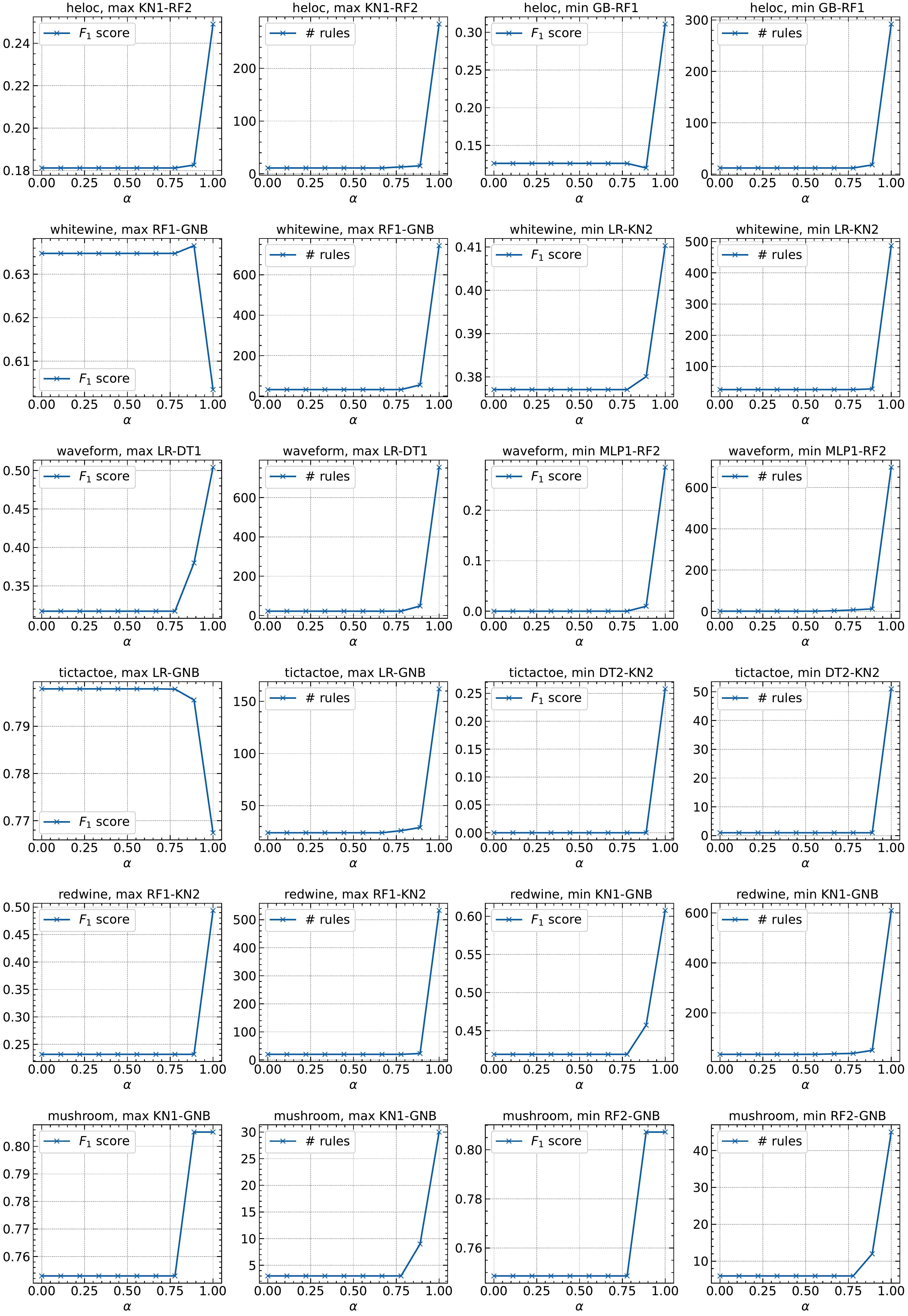}
    \caption{Effect of $\alpha$ on metrics for IMD. The algorithm is sensitive to $\alpha$ close to $1$ and quickly makes a transition from conjoined trees to completely separate trees.}
    \label{fig:metrics-vs-alpha1}
\end{figure}

In Figures~\ref{fig:metrics-vs-alpha} and ~\ref{fig:metrics-vs-alpha1}, we show this trend for each dataset, and max and min gap model pairs as used earlier (Section~\ref{sec:expt:sep}).

Almost in all the cases, we observe sharp increase when $\alpha$ gets close to 1 i.e., the surrogates diverge away from each other. The figures in y-axis also match with prior results for separate surrogates and IMD. 

\paragraph{Conclusion} The algorithm is sensitive to $\alpha$ close to $1$ and quickly makes a transition from conjoined trees to completely separate trees. But close to zero (typically less than $0.5$ as seen from the plots), the metrics are \emph{not} sensitive to alpha.

\subsection{Perturbation/change detection experiment}
\label{appdx:perturbation-expt}
In this experiment, we artificially induce a change in the dataset (following ~\cite{nair2021changed}) and see if our method faithfully recovers it.

\paragraph{Set up} For each dataset, on the available training data with ground truth labels we first fit a model, and call it the \emph{original} model. Then, we design a rule $r$, flip labels of all instances in the training data satisfying the rule with some probability $p$, and fit a \emph{revised} model with the same parameters as used in the \emph{original} model. We then subject these two models --- \emph{original} and \emph{revised} --- to model differencing (using proposed IMD method and separate surrogates) and verify if the extracted diff-rules recover the change inducing rule $r$.

For each dataset, and (original, revised) model combination, we vary the perturbation probability $p \in \{0.5, 0.6, \dots, 1.0\}$, and find the diff-rule having maximum similarity with the perturbing rule $r$ (using statistical and semantic rule similarity measures ~\citep{nair2021changed}). We consider the diff-rule to have recovered the perturbation if the similarity is greater than a threshold value ($\geq 0.3$ in the experiments). Note that this is a stricter definition of recovery than just covering the support of $r$, since covering may be achieved jointly by multiple diff rules with none of them having high similarity to 
on their own.
Also, even when we perturb the data with a single rule $r$ with some probability $p$, we observed that in the retrieved diff-ruleset, often there are multiple diff-rules that overlap (or have statistical or semantic similarity as mentioned above) with $r$.

We have experimented with 8 binary classification datasets (for simplicity in flipping class labels), 4 model classes (LR, DT2, RF1, GB) and for 5 random rules for each perturbation probability. We have also repeated the experiment for two methods of model differencing: our IMD method and separate surrogates approach (both with \texttt{max\_depth=6}, as used in prior experiments).

\begin{figure}[ph]
    \centering
    \includegraphics[width=0.9\textwidth]{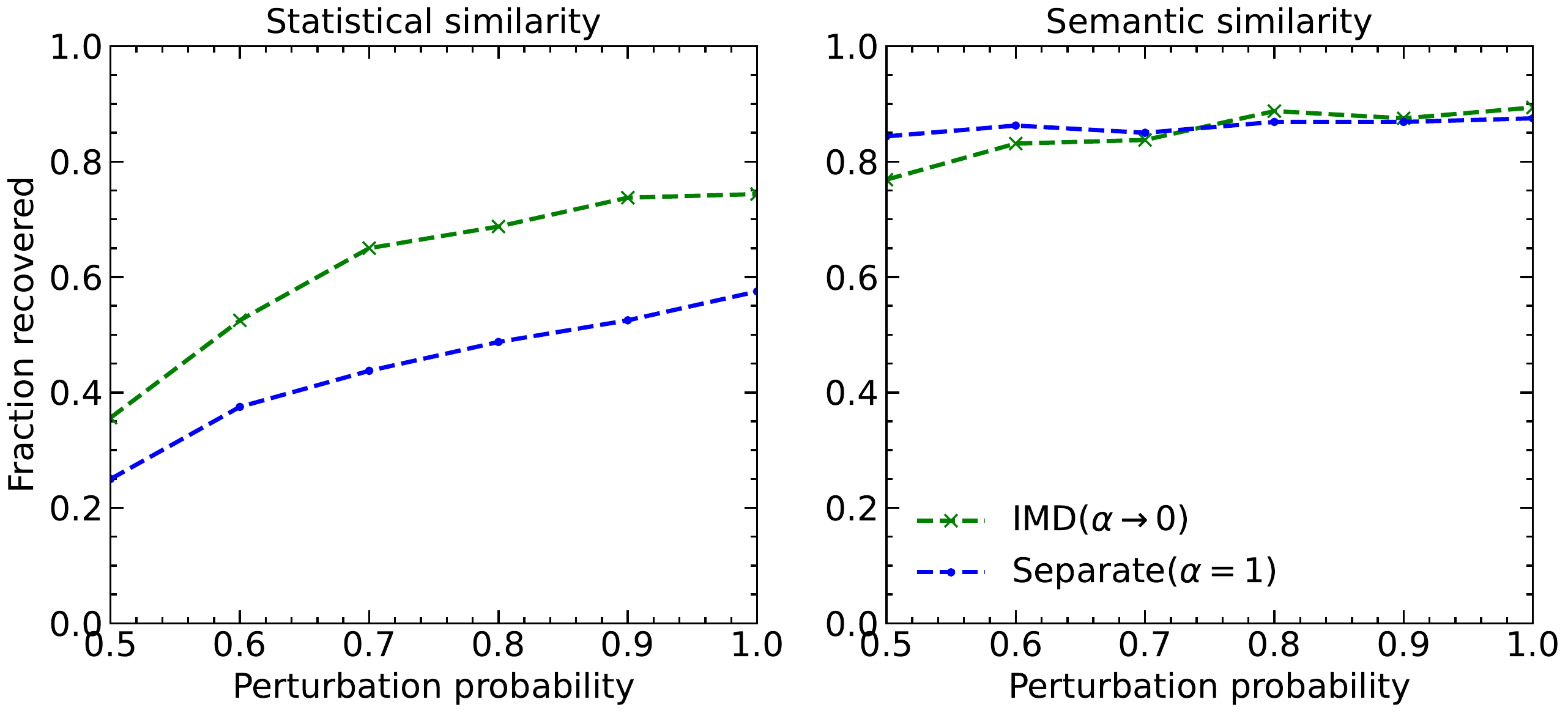}
    \caption{Both baselines are able to recover the perturbation (rule) with high similarity scores.}
    \label{fig:perturbation}
\end{figure}

We show the averaged plots in Figure~\ref{fig:perturbation}.

\paragraph{Conclusion} We see for both methods the recovery rate increases with perturbation probability, for both statistical and semantic similarity measures. Interestingly, IMD does better in recovering the rule than separate, because separate surrogates approach tends to generate many (and more granular) regions in the feature space to cover a larger region, whereas IMD is constrained by difference localization.

This also highlights that the differences are reliably identified by IMD in its current configuration.

%%%%%%%%%%%%%%%%%%%%%%%%%%%%%%%%%%%%%%%%%%
%%%%%%%%%%%%%%%%%%%%%%%%%%%%%%%%%%%%%%%%%%

\clearpage
\section{ADDITIONAL DETAILS ON CASE STUDY}
\label{appdx:casestudy}

In several domains involving people, machine learning models may systematically advantage certain cohorts and lead to unfair outcomes. This may be due to biases in the training data, mechanisms of data labeling and collection, or other factors. Several bias mitigation methods have been proposed to update models to correct for such biases. Our differencing method can be used to find differences in models before and after mitigation, to ensure that there were no knock-on effects of the change. 

Our example is from the advertising industry where targeted ads are personalized based on user profiles. Companies through ad campaigns want to reach potential customers and use machine learning models to predict who they might be. But what models predict and actual conversions, i.e. when some one interacts with an ad, can be very different and biased. This is a poor outcome for companies who are not reaching the right audience, and for customers who are incorrectly targeted.

Using synthetic data generated from an actual ad campaign, we train a model, detect (using Multi-dimensional subset scan (MDSS) \cite{zhang2016identifying}). For discovered privileged groups we measure bias using Disparate Impact Ratio, which is the likelihood of positive outcomes for unprivileged groups by positive outcomes for privileged group. A ratio of close to 1 indicates parity between two groups, while values away from 1 imply advantage to one of the groups. 

In our case, MDSS reported a privileged group of \emph{non}-homeowners, where predicted conversions were considerably higher than ground truth. Using this as the privileged group, the disparate impact ratio was found to be $0.48$, indicated favored status for this cohort. The test balanced accuracy of this base model is $0.5723$.

To remedy bias, we apply a mitigation using Reject Option Classification \citep{kamiran2012decision}, a post-processing technique that gives favorable outcomes to unprivileged groups and unfavorable outcomes to privileged groups in a confidence band around the decision boundary with the highest uncertainty
\footnote{\texttt{https://github.com/Trusted-AI/AIF360/blob/master/examples/tutorial\_bias\_advertising.ipynb}}
% \footnote{Anonymized link}
. After mitigation the disparate impact ratio improves to $1.1$ indicating better parity between groups with the test balanced accuracy of $0.568$, i.e. a negligible drop in predictive accuracy.

\begin{figure}[ht]
    \centering
    \includegraphics[width=\textwidth]{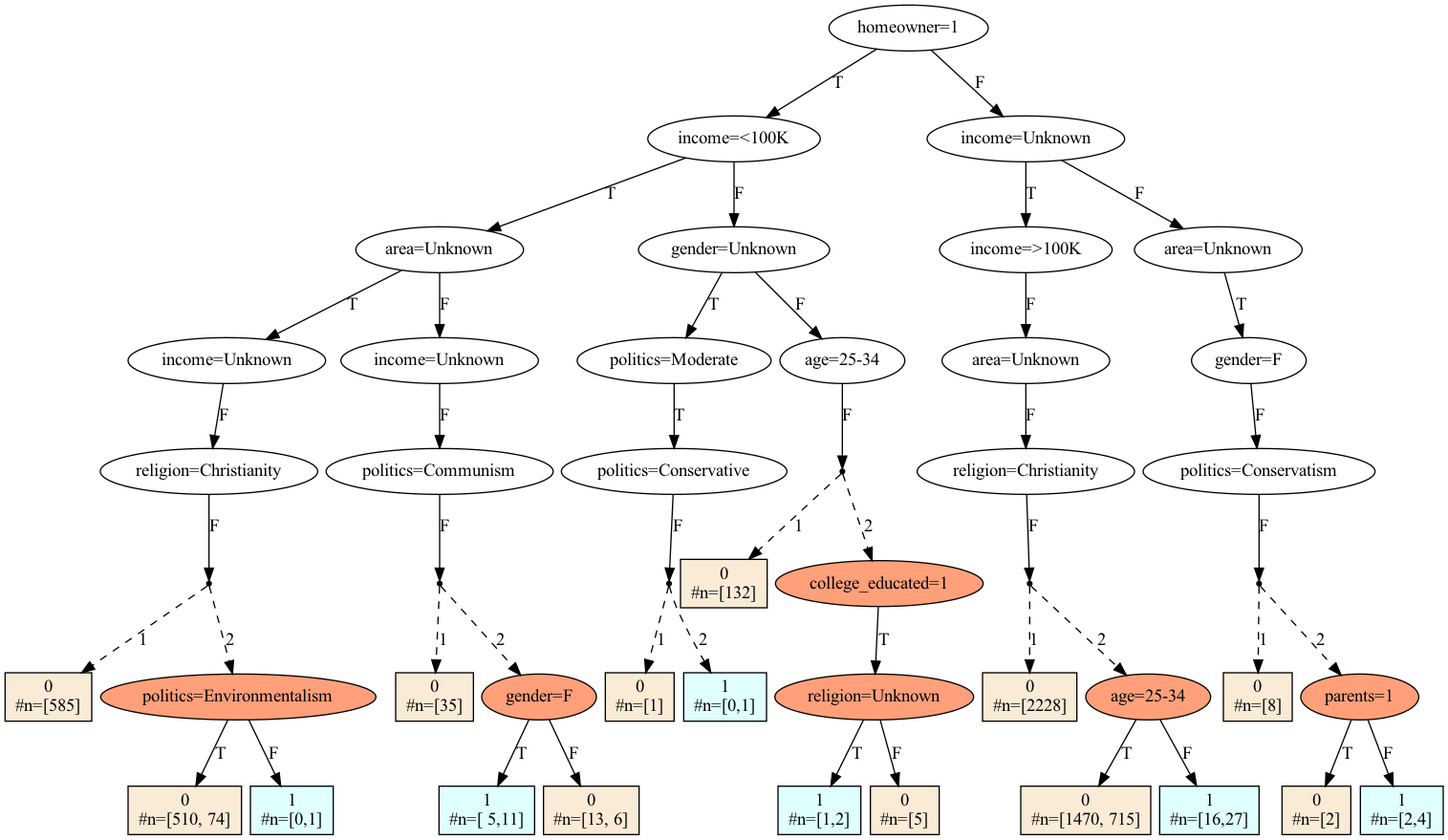}
    \caption{Joint Surrogate Tree for models before and after bias mitigation using Reject Option Classification. Parts of the tree where the outcome labels for both models are the same are hidden.}
    \label{fig:ad:fulltree}
\end{figure}

%%%%%%%%%%%%%%%%%%%%%%%%%%%%%%%%%%%%%%%%%%
%%%%%%%%%%%%%%%%%%%%%%%%%%%%%%%%%%%%%%%%%%

% additional tables

\clearpage
\section{ADDITIONAL TABLES}
\label{appdx: full-tables}
We have provided the complete versions of tables presented in the main paper with all the missing metrics, and standard deviations over 5 runs here.

\begin{sidewaystable}

\centering

\caption{Full table for Separate Surrogate vs.~IMD technique with added number of predicates (\#p) column, and standard deviation values over 5 runs. We removed the model pairs column (available in Table~\ref{tab:approaches-prre-std})
%main paper Table~\ref{tab:ablations-with-deltas-trimmed-for-uai}) 
for brevity.} 

\begin{tabular}{lccrrccrr}
\toprule
          &  \multicolumn{4}{c}{\textbf{Separate Surrogates}} & \multicolumn{4}{c}{\textbf{IMD}} \\
          
          \cmidrule(r){2-5} \cmidrule(l){6-9}

\textbf{Dataset}     &      Pr &      Re &      \#r &     \#p &             Pr &      Re &    \#r &    \#p \\
\midrule
\multirow{2}{*}{adult} &     0.96 $\pm$ 0.00 &   0.88 $\pm$ 0.00 &    70.0 $\pm$ 17.77 &    69.2 $\pm$ 2.99 &       0.96 $\pm$ 0.00 &   0.88 $\pm$ 0.00 &  18.0 $\pm$ 1.26 &   48.4 $\pm$ 3.01 \\
          &  0.45 $\pm$ 0.02 &  0.29 $\pm$ 0.13 &    155.4 $\pm$ 7.74 &    127.8 $\pm$ 7.7 &      0.46 $\pm$ 0.02 &  0.16 $\pm$ 0.01 &  17.4 $\pm$ 1.96 &   49.2 $\pm$ 5.08 \\
\cline{1-9}
\multirow{2}{*}{bankm} &   0.66 $\pm$ 0.01 &  0.75 $\pm$ 0.01 &   263.6 $\pm$ 20.03 &   172.0 $\pm$ 9.38 &       0.70 $\pm$ 0.04 &  0.67 $\pm$ 0.02 &  23.0 $\pm$ 2.76 &   66.8 $\pm$ 6.14 \\
          &  0.74 $\pm$ 0.02 &  0.75 $\pm$ 0.02 &    345.0 $\pm$ 7.16 &   177.0 $\pm$ 3.22 &      0.71 $\pm$ 0.01 &  0.69 $\pm$ 0.01 &  34.4 $\pm$ 1.85 &   83.2 $\pm$ 6.05 \\
\cline{1-9}
\multirow{2}{*}{banknote} &  0.88 $\pm$ 0.02 &  0.89 $\pm$ 0.05 &     32.2 $\pm$ 5.38 &    48.2 $\pm$ 5.91 &       0.90 $\pm$ 0.04 &  0.88 $\pm$ 0.04 &  13.4 $\pm$ 2.33 &   37.4 $\pm$ 5.39 \\
          &  0.53 $\pm$ 0.06 &   0.60 $\pm$ 0.04 &     30.8 $\pm$ 8.95 &     52.0 $\pm$ 7.4 &      0.64 $\pm$ 0.14 &  0.47 $\pm$ 0.21 &   7.2 $\pm$ 2.71 &   25.0 $\pm$ 6.51 \\
\cline{1-9}
\multirow{2}{*}{bc} &  0.38 $\pm$ 0.08 &  0.46 $\pm$ 0.09 &     39.2 $\pm$ 7.57 &    34.0 $\pm$ 2.83 &      0.44 $\pm$ 0.16 &   0.40 $\pm$ 0.17 &   9.6 $\pm$ 1.36 &    27.0 $\pm$ 2.0 \\
          &  0.37 $\pm$ 0.10 &  0.38 $\pm$ 0.11 &     49.0 $\pm$ 7.77 &     38.8 $\pm$ 2.4 &       0.30 $\pm$ 0.05 &  0.24 $\pm$ 0.15 &   10.8 $\pm$ 1.6 &   29.4 $\pm$ 2.94 \\
\cline{1-9}
\multirow{2}{*}{diabetes} &  0.42 $\pm$ 0.09 &  0.45 $\pm$ 0.06 &   215.8 $\pm$ 36.22 &   131.4 $\pm$ 7.53 &       0.40 $\pm$ 0.03 &  0.28 $\pm$ 0.07 &  24.2 $\pm$ 3.37 &   70.6 $\pm$ 7.34 \\
          &  0.39 $\pm$ 0.06 &  0.43 $\pm$ 0.03 &   156.0 $\pm$ 29.35 &   112.2 $\pm$ 8.18 &      0.31 $\pm$ 0.08 &  0.34 $\pm$ 0.13 &  20.8 $\pm$ 2.04 &   56.6 $\pm$ 2.58 \\
\cline{1-9}
\multirow{2}{*}{eye} & 0.65 $\pm$ 0.02 &  0.66 $\pm$ 0.08 &  1054.0 $\pm$ 57.48 &   175.6 $\pm$ 7.42 &       0.60 $\pm$ 0.04 &  0.71 $\pm$ 0.18 &   36.2 $\pm$ 2.4 &    86.4 $\pm$ 5.2 \\
          &  0.59 $\pm$ 0.01 &  0.53 $\pm$ 0.02 &   781.6 $\pm$ 32.22 &   227.2 $\pm$ 6.27 &      0.57 $\pm$ 0.02 &  0.39 $\pm$ 0.05 &  28.4 $\pm$ 1.36 &   76.6 $\pm$ 1.74 \\
\cline{1-9}
\multirow{2}{*}{heloc} &  0.40 $\pm$ 0.03 &  0.23 $\pm$ 0.04 &   373.0 $\pm$ 58.53 &   200.0 $\pm$ 4.56 &       0.40 $\pm$ 0.05 &  0.13 $\pm$ 0.05 &  15.8 $\pm$ 3.87 &   53.8 $\pm$ 8.93 \\
          &  0.30 $\pm$ 0.02 &  0.19 $\pm$ 0.08 &   234.4 $\pm$ 45.63 &   186.8 $\pm$ 4.53 &      0.25 $\pm$ 0.04 &  0.06 $\pm$ 0.02 &   14.6 $\pm$ 1.2 &   47.2 $\pm$ 4.17 \\
\cline{1-9}
\multirow{2}{*}{magic} &  0.75 $\pm$ 0.02 &  0.58 $\pm$ 0.05 &   362.8 $\pm$ 29.31 &   179.2 $\pm$ 5.78 &      0.75 $\pm$ 0.02 &  0.52 $\pm$ 0.04 &   25.0 $\pm$ 1.1 &   71.0 $\pm$ 3.58 \\
          &  0.43 $\pm$ 0.04 &  0.36 $\pm$ 0.02 &    282.6 $\pm$ 9.39 &   217.4 $\pm$ 2.33 &      0.42 $\pm$ 0.05 &  0.17 $\pm$ 0.04 &   11.0 $\pm$ 1.1 &   41.8 $\pm$ 3.97 \\
\cline{1-9}
\multirow{2}{*}{mushroom} &  0.94 $\pm$ 0.02 &   0.70 $\pm$ 0.02 &     28.0 $\pm$ 4.86 &     30.6 $\pm$ 1.5 &      0.81 $\pm$ 0.08 &   0.70 $\pm$ 0.03 &    5.0 $\pm$ 0.0 &   16.0 $\pm$ 0.63 \\
          & 0.93 $\pm$ 0.03 &   0.70 $\pm$ 0.02 &      42.0 $\pm$ 1.9 &     38.6 $\pm$ 1.5 &      0.74 $\pm$ 0.07 &  0.71 $\pm$ 0.02 &    8.6 $\pm$ 1.5 &   23.2 $\pm$ 1.72 \\
\cline{1-9}
\multirow{2}{*}{redwine} &  0.46 $\pm$ 0.03 &  0.52 $\pm$ 0.02 &   627.8 $\pm$ 64.99 &   189.8 $\pm$ 6.52 &      0.52 $\pm$ 0.06 &  0.25 $\pm$ 0.05 &  29.0 $\pm$ 1.41 &   78.2 $\pm$ 3.87 \\
          &   0.70 $\pm$ 0.04 &  0.59 $\pm$ 0.07 &   563.6 $\pm$ 89.88 &  175.0 $\pm$ 12.85 &      0.69 $\pm$ 0.05 &   0.47 $\pm$ 0.10 &   40.4 $\pm$ 4.5 &   93.8 $\pm$ 5.71 \\
\cline{1-9}
\multirow{2}{*}{tictactoe} & 0.76 $\pm$ 0.03 &  0.78 $\pm$ 0.04 &    109.6 $\pm$ 26.7 &     42.4 $\pm$ 1.5 &      0.76 $\pm$ 0.02 &  0.89 $\pm$ 0.05 &  24.4 $\pm$ 3.01 &   35.0 $\pm$ 1.41 \\
          &  0.10 $\pm$ 0.08 &  0.15 $\pm$ 0.09 &     54.0 $\pm$ 33.7 &    38.8 $\pm$ 2.64 &      0.16 $\pm$ 0.14 &  0.11 $\pm$ 0.09 &    5.8 $\pm$ 2.4 &   21.2 $\pm$ 6.49 \\
\cline{1-9}
\multirow{2}{*}{waveform} & 0.45 $\pm$ 0.04 &  0.52 $\pm$ 0.02 &   746.0 $\pm$ 43.74 &   199.2 $\pm$ 0.98 &      0.49 $\pm$ 0.02 &  0.27 $\pm$ 0.04 &  33.2 $\pm$ 7.73 &  95.8 $\pm$ 18.73 \\
          &  0.17 $\pm$ 0.04 &  0.32 $\pm$ 0.03 &    725.0 $\pm$ 91.1 &   237.2 $\pm$ 7.52 &       0.10 $\pm$ 0.05 &  0.02 $\pm$ 0.01 &   9.0 $\pm$ 3.52 &  33.6 $\pm$ 11.71 \\
\cline{1-9}
\multirow{2}{*}{whitewine} &  0.64 $\pm$ 0.02 &  0.59 $\pm$ 0.03 &  847.2 $\pm$ 113.12 &    219.2 $\pm$ 7.0 &      0.63 $\pm$ 0.01 &  0.56 $\pm$ 0.08 &  42.6 $\pm$ 4.27 &   99.6 $\pm$ 6.71 \\
          &  0.56 $\pm$ 0.03 &  0.33 $\pm$ 0.02 &    580.0 $\pm$ 63.9 &   201.2 $\pm$ 6.11 &      0.55 $\pm$ 0.02 &  0.35 $\pm$ 0.04 &  36.6 $\pm$ 4.67 &   92.0 $\pm$ 6.54 \\
\bottomrule
\end{tabular}

\label{tab:ablations-std}
\end{sidewaystable}

\begin{table*}[h]
\centering

\caption{Comparison of F1-scores with standard deviation over 5 runs. Means are already reported in main paper Table~\ref{tab:approaches}.}

\begin{tabular}{lrrrrr}
\toprule
% \hline
          &   & \textbf{Sep.} & \textbf{Direct} & \textbf{Direct} & \textbf{BRCG} \\
          
\textbf{Dataset} & \textbf{IMD} &  \textbf{Surr.} &      \textbf{DT} &        \textbf{GB} &             \textbf{Diff.} \\

\midrule

\multirow{2}{*}{adult} &   0.92 $\pm$ 0.00 &     0.92 $\pm$ 0.00 &   0.92 $\pm$ 0.00 &   0.98 $\pm$ 0.00 &   0.33 $\pm$ 0.01 \\
          &     0.23 $\pm$ 0.01 &    0.34 $\pm$ 0.09 &  0.17 $\pm$ 0.11 &  0.61 $\pm$ 0.01 &   0.31 $\pm$ 0.01 \\
\cline{1-6}
\multirow{2}{*}{bankm} &     0.68 $\pm$ 0.02 &      0.70 $\pm$ 0.00 &  0.69 $\pm$ 0.02 &  0.77 $\pm$ 0.01 &   0.41 $\pm$ 0.01 \\
          &      0.70 $\pm$ 0.00 &    0.75 $\pm$ 0.01 &  0.68 $\pm$ 0.01 &  0.82 $\pm$ 0.01 &    0.41 $\pm$ 0.00 \\
\cline{1-6}
\multirow{2}{*}{banknote} &    0.89 $\pm$ 0.02 &    0.89 $\pm$ 0.02 &  0.83 $\pm$ 0.02 &  0.94 $\pm$ 0.01 &   0.27 $\pm$ 0.01 \\
          &     0.52 $\pm$ 0.17 &    0.56 $\pm$ 0.04 &  0.57 $\pm$ 0.09 &  0.63 $\pm$ 0.06 &   0.06 $\pm$ 0.02 \\
\cline{1-6}
\multirow{2}{*}{bc} &    0.39 $\pm$ 0.16 &    0.41 $\pm$ 0.06 &  0.17 $\pm$ 0.13 &    0.00 $\pm$ 0.00 &    0.10 $\pm$ 0.02 \\
          &    0.25 $\pm$ 0.11 &    0.37 $\pm$ 0.11 &  0.28 $\pm$ 0.12 &  0.19 $\pm$ 0.12 &   0.13 $\pm$ 0.03 \\
\cline{1-6}
\multirow{2}{*}{diabetes} &     0.32 $\pm$ 0.04 &    0.43 $\pm$ 0.08 &  0.21 $\pm$ 0.04 &  0.35 $\pm$ 0.05 &   0.35 $\pm$ 0.01 \\
          &    0.32 $\pm$ 0.09 &    0.41 $\pm$ 0.04 &  0.09 $\pm$ 0.07 &  0.22 $\pm$ 0.05 &    0.30 $\pm$ 0.07 \\
\cline{1-6}
\multirow{2}{*}{eye} &     0.63 $\pm$ 0.09 &    0.65 $\pm$ 0.02 &  0.72 $\pm$ 0.01 &  0.74 $\pm$ 0.01 &     --- \\
          &     0.46 $\pm$ 0.04 &    0.56 $\pm$ 0.01 &  0.48 $\pm$ 0.03 &  0.59 $\pm$ 0.01 &     --- \\
\cline{1-6}
\multirow{2}{*}{heloc} &    0.19 $\pm$ 0.05 &    0.29 $\pm$ 0.03 &  0.03 $\pm$ 0.01 &  0.14 $\pm$ 0.02 &   0.37 $\pm$ 0.02 \\
          &     0.10 $\pm$ 0.02 &    0.22 $\pm$ 0.04 &  0.02 $\pm$ 0.01 &  0.05 $\pm$ 0.03 &   0.27 $\pm$ 0.07 \\
\cline{1-6}
\multirow{2}{*}{magic} &     0.62 $\pm$ 0.02 &    0.65 $\pm$ 0.03 &  0.63 $\pm$ 0.04 &  0.78 $\pm$ 0.04 &    0.40 $\pm$ 0.02 \\
          &    0.24 $\pm$ 0.03 &    0.39 $\pm$ 0.01 &  0.14 $\pm$ 0.03 &  0.27 $\pm$ 0.02 &    0.20 $\pm$ 0.02 \\
\cline{1-6}
\multirow{2}{*}{mushroom} &    0.75 $\pm$ 0.05 &     0.80 $\pm$ 0.02 &  0.81 $\pm$ 0.05 &  0.97 $\pm$ 0.02 &   0.76 $\pm$ 0.03 \\
          &   0.72 $\pm$ 0.04 &     0.80 $\pm$ 0.02 &  0.81 $\pm$ 0.02 &  0.97 $\pm$ 0.02 &   0.74 $\pm$ 0.04 \\
\cline{1-6}
\multirow{2}{*}{redwine} &    0.33 $\pm$ 0.05 &    0.49 $\pm$ 0.02 &  0.17 $\pm$ 0.07 &  0.48 $\pm$ 0.02 &     --- \\
          &    0.55 $\pm$ 0.07 &    0.63 $\pm$ 0.03 &  0.64 $\pm$ 0.06 &  0.75 $\pm$ 0.02 &     --- \\
\cline{1-6}
\multirow{2}{*}{tictactoe} &    0.82 $\pm$ 0.02 &    0.77 $\pm$ 0.04 &  0.77 $\pm$ 0.01 &  0.82 $\pm$ 0.02 &   0.83 $\pm$ 0.04 \\
          &   0.12 $\pm$ 0.10 &    0.12 $\pm$ 0.08 &    0.00 $\pm$ 0.00 &  0.09 $\pm$ 0.08 &     0.00 $\pm$ 0.00 \\
\cline{1-6}
\multirow{2}{*}{waveform} &   0.34 $\pm$ 0.04 &    0.48 $\pm$ 0.02 &  0.15 $\pm$ 0.06 &  0.13 $\pm$ 0.03 &     --- \\
          &   0.04 $\pm$ 0.02 &    0.22 $\pm$ 0.04 &  0.07 $\pm$ 0.01 &  0.06 $\pm$ 0.01 &     --- \\
\cline{1-6}
\multirow{2}{*}{whitewine} &   0.59 $\pm$ 0.04 &    0.61 $\pm$ 0.01 &  0.62 $\pm$ 0.03 &  0.74 $\pm$ 0.02 &     --- \\
          &   0.43 $\pm$ 0.04 &    0.41 $\pm$ 0.02 &  0.43 $\pm$ 0.06 &  0.64 $\pm$ 0.01 &     --- \\
\bottomrule

\end{tabular}

\label{tab:approaches-f1-std}
\end{table*}

\begin{table*}[h]
\centering

\caption{Precision values with standard deviation over 5 runs. Means are already reported in main paper Table~\ref{tab:refinement} for some of the benchmarks.}

\begin{tabular}{lccc}
\toprule
% \hline
    \textbf{Dataset}     & $\text{\bf IMD}_\text{\bf6}$ & $\text{\bf IMD}_\text{\bf6+1}$ & $\text{\bf IMD}_\text{\bf7}$ \\

\midrule

\multirow{2}{*}{adult} &       0.96 $\pm$ 0.00 &             0.96 $\pm$ 0.00 &      0.95 $\pm$ 0.01 \\
          &    0.46 $\pm$ 0.02 &            0.59 $\pm$ 0.03 &      0.53 $\pm$ 0.05 \\
\cline{1-4}
\multirow{2}{*}{bankm} &      0.70 $\pm$ 0.04 &            0.78 $\pm$ 0.02 &      0.77 $\pm$ 0.03 \\
          &    0.71 $\pm$ 0.01 &            0.79 $\pm$ 0.02 &      0.74 $\pm$ 0.02 \\
\cline{1-4}
\multirow{2}{*}{banknote} &      0.90 $\pm$ 0.04 &             0.90 $\pm$ 0.03 &      0.89 $\pm$ 0.01 \\
          &    0.64 $\pm$ 0.14 &            0.67 $\pm$ 0.15 &      0.76 $\pm$ 0.06 \\
\cline{1-4}
\multirow{2}{*}{bc} &    0.44 $\pm$ 0.16 &            0.44 $\pm$ 0.16 &      0.44 $\pm$ 0.16 \\
          &     0.30 $\pm$ 0.05 &            0.28 $\pm$ 0.08 &      0.26 $\pm$ 0.08 \\
\cline{1-4}
\multirow{2}{*}{diabetes} &     0.40 $\pm$ 0.03 &            0.47 $\pm$ 0.05 &      0.44 $\pm$ 0.07 \\
          &    0.31 $\pm$ 0.08 &            0.33 $\pm$ 0.11 &      0.34 $\pm$ 0.06 \\
\cline{1-4}
\multirow{2}{*}{eye} &     0.60 $\pm$ 0.04 &            0.67 $\pm$ 0.01 &      0.62 $\pm$ 0.04 \\
          &     0.57 $\pm$ 0.02 &            0.64 $\pm$ 0.04 &      0.57 $\pm$ 0.02 \\
\cline{1-4}
\multirow{2}{*}{heloc} &      0.40 $\pm$ 0.05 &            0.45 $\pm$ 0.06 &      0.42 $\pm$ 0.03 \\
          &    0.25 $\pm$ 0.04 &            0.25 $\pm$ 0.02 &      0.26 $\pm$ 0.02 \\
\cline{1-4}
\multirow{2}{*}{magic} &    0.75 $\pm$ 0.02 &             0.80 $\pm$ 0.01 &      0.73 $\pm$ 0.02 \\
          &    0.42 $\pm$ 0.05 &            0.55 $\pm$ 0.05 &      0.46 $\pm$ 0.05 \\
\cline{1-4}
\multirow{2}{*}{mushroom} &     0.81 $\pm$ 0.08 &            0.95 $\pm$ 0.03 &      0.88 $\pm$ 0.05 \\
          &    0.74 $\pm$ 0.07 &            0.94 $\pm$ 0.04 &      0.89 $\pm$ 0.06 \\
\cline{1-4}
\multirow{2}{*}{redwine} &     0.52 $\pm$ 0.06 &            0.56 $\pm$ 0.07 &      0.48 $\pm$ 0.03 \\
          &   0.69 $\pm$ 0.05 &            0.73 $\pm$ 0.06 &      0.68 $\pm$ 0.04 \\
\cline{1-4}
\multirow{2}{*}{tictactoe} &     0.76 $\pm$ 0.02 &            0.79 $\pm$ 0.03 &      0.78 $\pm$ 0.03 \\
          &    0.16 $\pm$ 0.14 &            0.19 $\pm$ 0.19 &      0.18 $\pm$ 0.14 \\
\cline{1-4}
\multirow{2}{*}{waveform} &    0.49 $\pm$ 0.02 &            0.54 $\pm$ 0.03 &      0.49 $\pm$ 0.01 \\
          &      0.10 $\pm$ 0.05 &            0.14 $\pm$ 0.04 &      0.17 $\pm$ 0.06 \\
\cline{1-4}
\multirow{2}{*}{whitewine} &    0.63 $\pm$ 0.01 &            0.67 $\pm$ 0.03 &      0.64 $\pm$ 0.01 \\
          &    0.55 $\pm$ 0.02 &            0.59 $\pm$ 0.03 &      0.59 $\pm$ 0.02 \\
\bottomrule
\end{tabular}

\label{tab:refinement-std}
\end{table*}

\begin{sidewaystable}[h]
% \scriptsize
% \small
\centering

\caption{Precision, Recall and Interpretability metrics for $\textrm{IMD}_{6+1}$ and $\textrm{IMD}_{7}$. The values for $\textrm{IMD}_{6}$ are reported earlier in Table~\ref{tab:ablations-std}. As can be seen, the recall values for $\textrm{IMD}_{6+1}$ are consistently lower than both $\textrm{IMD}_{6}$ and $\textrm{IMD}_{7}$. Also, as expected, $\textrm{IMD}_{6+1}$ has lower no. of rules and no. of predicates than $\textrm{IMD}_{7}$ since it does selective splitting of nodes.
}

\begin{tabular}{lccrrccrr}
\toprule
          &  \multicolumn{4}{c}{$\text{\bf IMD}_\text{\bf6+1}$} & \multicolumn{4}{c}{$\text{\bf IMD}_\text{\bf7}$} \\
          
          \cmidrule(r){2-5} \cmidrule(l){6-9}
\textbf{D}     &      Pr &      Re &      \#r &     \#p &             Pr &      Re &    \#r &    \#p \\
\midrule
\multirow{2}{*}{adult} &         0.96 $\pm$ 0.00 &   0.88 $\pm$ 0.00 &   18.0 $\pm$ 1.26 &    51.8 $\pm$ 3.31 &      0.95 $\pm$ 0.01 &   0.90 $\pm$ 0.01 &   29.4 $\pm$ 3.98 &    72.0 $\pm$ 7.07 \\
          &          0.59 $\pm$ 0.03 &   0.12 $\pm$ 0.00 &   22.2 $\pm$ 3.31 &    68.6 $\pm$ 7.26 &      0.53 $\pm$ 0.05 &  0.14 $\pm$ 0.06 &   36.8 $\pm$ 3.12 &    97.8 $\pm$ 6.21 \\
\cline{1-9}
\multirow{2}{*}{bankm} &        0.78 $\pm$ 0.02 &  0.61 $\pm$ 0.03 &   26.4 $\pm$ 3.72 &    84.6 $\pm$ 7.55 &      0.77 $\pm$ 0.03 &  0.64 $\pm$ 0.02 &    45.0 $\pm$ 2.0 &   124.2 $\pm$ 5.91 \\
          &      0.79 $\pm$ 0.02 &  0.62 $\pm$ 0.01 &   40.2 $\pm$ 4.07 &   108.8 $\pm$ 7.36 &      0.74 $\pm$ 0.02 &  0.71 $\pm$ 0.03 &   58.6 $\pm$ 2.24 &    143.0 $\pm$ 3.9 \\
\cline{1-9}
\multirow{2}{*}{banknote} &        0.90 $\pm$ 0.03 &  0.89 $\pm$ 0.04 &   13.8 $\pm$ 2.64 &    38.8 $\pm$ 6.11 &      0.89 $\pm$ 0.01 &  0.91 $\pm$ 0.03 &   14.8 $\pm$ 1.94 &    42.4 $\pm$ 3.93 \\
          &       0.67 $\pm$ 0.15 &  0.48 $\pm$ 0.21 &    7.8 $\pm$ 3.25 &    26.6 $\pm$ 7.61 &      0.76 $\pm$ 0.06 &  0.65 $\pm$ 0.09 &    8.4 $\pm$ 2.87 &    28.2 $\pm$ 6.34 \\
\cline{1-9}
\multirow{2}{*}{bc} &      0.44 $\pm$ 0.16 &   0.40 $\pm$ 0.17 &    9.8 $\pm$ 1.17 &    27.8 $\pm$ 1.33 &      0.44 $\pm$ 0.16 &   0.40 $\pm$ 0.17 &    9.6 $\pm$ 1.36 &    27.2 $\pm$ 1.72 \\
          &       0.28 $\pm$ 0.08 &  0.22 $\pm$ 0.16 &   11.4 $\pm$ 1.85 &    30.6 $\pm$ 3.07 &      0.26 $\pm$ 0.08 &  0.22 $\pm$ 0.16 &   11.0 $\pm$ 1.41 &    30.0 $\pm$ 2.83 \\
\cline{1-9}
\multirow{2}{*}{diabetes} &       0.47 $\pm$ 0.05 &  0.27 $\pm$ 0.04 &   28.0 $\pm$ 5.66 &   82.6 $\pm$ 12.71 &      0.44 $\pm$ 0.07 &  0.39 $\pm$ 0.04 &   36.6 $\pm$ 4.13 &   102.6 $\pm$ 9.02 \\
          &        0.33 $\pm$ 0.11 &   0.26 $\pm$ 0.10 &   23.2 $\pm$ 3.31 &     69.2 $\pm$ 6.4 &      0.34 $\pm$ 0.06 &  0.36 $\pm$ 0.09 &   30.0 $\pm$ 1.67 &    84.8 $\pm$ 1.72 \\
\cline{1-9}
\multirow{2}{*}{eye} &       0.67 $\pm$ 0.01 &  0.49 $\pm$ 0.09 &    66.0 $\pm$ 2.0 &   152.4 $\pm$ 5.54 &      0.62 $\pm$ 0.04 &  0.69 $\pm$ 0.17 &    72.6 $\pm$ 2.8 &   167.6 $\pm$ 7.12 \\
          &        0.64 $\pm$ 0.04 &  0.31 $\pm$ 0.05 &   43.8 $\pm$ 4.45 &   123.8 $\pm$ 9.74 &      0.57 $\pm$ 0.02 &  0.47 $\pm$ 0.03 &   66.4 $\pm$ 5.16 &   168.2 $\pm$ 11.3 \\
\cline{1-9}
\multirow{2}{*}{heloc} &        0.45 $\pm$ 0.06 &  0.08 $\pm$ 0.03 &   20.2 $\pm$ 3.87 &   76.4 $\pm$ 12.19 &      0.42 $\pm$ 0.03 &  0.14 $\pm$ 0.02 &   39.8 $\pm$ 2.93 &   126.4 $\pm$ 5.16 \\
          &      0.25 $\pm$ 0.02 &  0.04 $\pm$ 0.01 &   21.2 $\pm$ 3.82 &   73.2 $\pm$ 11.65 &      0.26 $\pm$ 0.02 &   0.10 $\pm$ 0.04 &   38.4 $\pm$ 3.56 &   116.0 $\pm$ 8.79 \\
\cline{1-9}
\multirow{2}{*}{magic} &        0.80 $\pm$ 0.01 &   0.50 $\pm$ 0.04 &   28.4 $\pm$ 1.02 &    95.4 $\pm$ 6.65 &      0.73 $\pm$ 0.02 &  0.59 $\pm$ 0.03 &   50.6 $\pm$ 6.62 &  144.0 $\pm$ 14.79 \\
          &       0.55 $\pm$ 0.05 &  0.13 $\pm$ 0.02 &   14.6 $\pm$ 2.15 &    59.6 $\pm$ 6.22 &      0.46 $\pm$ 0.05 &  0.24 $\pm$ 0.05 &   32.2 $\pm$ 4.87 &  110.0 $\pm$ 15.79 \\
\cline{1-9}
\multirow{2}{*}{mushroom} &        0.95 $\pm$ 0.03 &   0.70 $\pm$ 0.03 &     6.0 $\pm$ 0.0 &     18.2 $\pm$ 0.4 &      0.88 $\pm$ 0.05 &  0.81 $\pm$ 0.07 &     7.6 $\pm$ 0.8 &    22.2 $\pm$ 2.14 \\
          &         0.94 $\pm$ 0.04 &  0.71 $\pm$ 0.02 &   10.2 $\pm$ 1.94 &    26.4 $\pm$ 2.24 &      0.89 $\pm$ 0.06 &  0.77 $\pm$ 0.08 &   13.2 $\pm$ 2.23 &    31.0 $\pm$ 0.63 \\
\cline{1-9}
\multirow{2}{*}{redwine} &         0.56 $\pm$ 0.07 &  0.17 $\pm$ 0.01 &   45.4 $\pm$ 5.08 &   112.0 $\pm$ 10.0 &      0.48 $\pm$ 0.03 &  0.33 $\pm$ 0.08 &   60.4 $\pm$ 7.17 &  142.8 $\pm$ 13.95 \\
          &         0.73 $\pm$ 0.06 &  0.41 $\pm$ 0.08 &   66.0 $\pm$ 2.83 &   145.4 $\pm$ 4.72 &      0.68 $\pm$ 0.04 &   0.56 $\pm$ 0.10 &   80.2 $\pm$ 3.54 &   173.6 $\pm$ 8.21 \\
\cline{1-9}
\multirow{2}{*}{tictactoe} &        0.79 $\pm$ 0.03 &  0.76 $\pm$ 0.05 &   26.8 $\pm$ 4.26 &     40.4 $\pm$ 0.8 &      0.78 $\pm$ 0.03 &   0.80 $\pm$ 0.06 &   32.6 $\pm$ 2.24 &    41.2 $\pm$ 1.17 \\
          &         0.19 $\pm$ 0.19 &  0.07 $\pm$ 0.06 &    6.4 $\pm$ 2.87 &    22.4 $\pm$ 7.26 &      0.18 $\pm$ 0.14 &  0.13 $\pm$ 0.05 &   13.0 $\pm$ 1.26 &    32.6 $\pm$ 3.44 \\
\cline{1-9}
\multirow{2}{*}{waveform} &        0.54 $\pm$ 0.03 &  0.24 $\pm$ 0.04 &  46.6 $\pm$ 11.76 &  128.4 $\pm$ 24.43 &      0.49 $\pm$ 0.01 &  0.39 $\pm$ 0.05 &  72.2 $\pm$ 11.03 &  188.2 $\pm$ 18.91 \\
          &         0.14 $\pm$ 0.04 &  0.02 $\pm$ 0.01 &   13.6 $\pm$ 5.82 &   48.4 $\pm$ 18.64 &      0.17 $\pm$ 0.06 &  0.08 $\pm$ 0.03 &   36.0 $\pm$ 2.97 &   123.4 $\pm$ 8.64 \\
\cline{1-9}
\multirow{2}{*}{whitewine} &     0.67 $\pm$ 0.03 &  0.45 $\pm$ 0.08 &   74.8 $\pm$ 8.47 &  169.6 $\pm$ 14.72 &      0.64 $\pm$ 0.01 &  0.54 $\pm$ 0.07 &  91.2 $\pm$ 11.74 &  200.0 $\pm$ 19.15 \\
          &         0.59 $\pm$ 0.03 &  0.28 $\pm$ 0.05 &  57.2 $\pm$ 11.02 &  144.0 $\pm$ 16.77 &      0.59 $\pm$ 0.02 &  0.37 $\pm$ 0.04 &  79.8 $\pm$ 10.17 &  190.2 $\pm$ 16.68 \\
\bottomrule
\end{tabular}

\label{tab:refinement-std-all-metrics}
\end{sidewaystable}

\begin{table*}[h]
% \small
\centering

\caption{Precision \& Recall values for Direct DT, Direct GB and BRCG Diff. with standard deviation over 5 runs. F1-scores are already reported in main paper Table~\ref{tab:approaches}.}

\begin{tabular}{llcccccc}
\toprule
          &            & \multicolumn{2}{c}{\textbf{Direct DT}} & \multicolumn{2}{c}{\textbf{Direct GB}} & \multicolumn{2}{c}{\textbf{BRCG Diff.}} \\

        \cmidrule(r){3-4} \cmidrule(l){5-6} \cmidrule(l){7-8}

\textbf{Dataset}      &   \textbf{$M_1$ vs. $M_2$}         &           Pr &           Re &           Pr &           Re &            Pr &           Re \\
\midrule
\multirow{2}{*}{adult} & max MLP1-GB &   0.96 $\pm$ 0.00 &   0.88 $\pm$ 0.00 &   0.98 $\pm$ 0.00 &   0.97 $\pm$ 0.00 &    0.20 $\pm$ 0.01 &    1.00 $\pm$ 0.00 \\
          & min MLP2-DT2 &  0.61 $\pm$ 0.06 &  0.11 $\pm$ 0.09 &  0.83 $\pm$ 0.01 &  0.48 $\pm$ 0.01 &   0.22 $\pm$ 0.02 &   0.59 $\pm$ 0.10 \\
\cline{1-8}
\multirow{2}{*}{bankm} & max MLP2-GB &  0.76 $\pm$ 0.02 &  0.63 $\pm$ 0.02 &  0.85 $\pm$ 0.02 &   0.70 $\pm$ 0.01 &   0.26 $\pm$ 0.01 &    1.00 $\pm$ 0.00 \\
          & min MLP1-GNB &  0.67 $\pm$ 0.01 &  0.69 $\pm$ 0.02 &  0.84 $\pm$ 0.01 &   0.80 $\pm$ 0.02 &    0.26 $\pm$ 0.00 &    1.00 $\pm$ 0.00 \\
\cline{1-8}
\multirow{2}{*}{banknote} & max KN1-GNB &  0.95 $\pm$ 0.04 &  0.74 $\pm$ 0.06 &  0.95 $\pm$ 0.02 &  0.92 $\pm$ 0.03 &   0.16 $\pm$ 0.01 &   1.00 $\pm$ 0.01 \\
          & min LR-DT1 &  0.69 $\pm$ 0.15 &   0.50 $\pm$ 0.09 &   0.80 $\pm$ 0.07 &  0.53 $\pm$ 0.09 &   0.03 $\pm$ 0.01 &    1.00 $\pm$ 0.00 \\
\cline{1-8}
\multirow{2}{*}{bc} & max DT1-GNB &   0.20 $\pm$ 0.13 &  0.16 $\pm$ 0.14 &    0.00 $\pm$ 0.00 &    0.00 $\pm$ 0.00 &   0.05 $\pm$ 0.01 &  0.86 $\pm$ 0.09 \\
          & min KN2-RF2 &   0.40 $\pm$ 0.24 &  0.23 $\pm$ 0.09 &  0.23 $\pm$ 0.12 &  0.17 $\pm$ 0.13 &   0.07 $\pm$ 0.02 &  0.89 $\pm$ 0.05 \\
\cline{1-8}
\multirow{2}{*}{diabetes} & max MLP2-GB &  0.43 $\pm$ 0.09 &  0.14 $\pm$ 0.03 &  0.57 $\pm$ 0.09 &  0.26 $\pm$ 0.03 &   0.22 $\pm$ 0.01 &  0.96 $\pm$ 0.03 \\
          & min RF1-GNB &  0.27 $\pm$ 0.21 &  0.08 $\pm$ 0.09 &   0.34 $\pm$ 0.10 &  0.17 $\pm$ 0.04 &   0.18 $\pm$ 0.04 &    1.00 $\pm$ 0.00 \\
\cline{1-8}
\multirow{2}{*}{eye} & max RF1-GNB &  0.61 $\pm$ 0.01 &   0.90 $\pm$ 0.01 &  0.68 $\pm$ 0.01 &  0.82 $\pm$ 0.03 &     --- &    --- \\
          & min LR-MLP1 &  0.62 $\pm$ 0.02 &   0.40 $\pm$ 0.04 &   0.70 $\pm$ 0.02 &  0.51 $\pm$ 0.01 &     --- &    --- \\
\cline{1-8}
\multirow{2}{*}{heloc} & max KN1-RF2 &  0.36 $\pm$ 0.08 &  0.02 $\pm$ 0.01 &  0.42 $\pm$ 0.01 &  0.09 $\pm$ 0.01 &   0.23 $\pm$ 0.01 &    1.00 $\pm$ 0.00 \\
          & min GB-RF1 &  0.16 $\pm$ 0.07 &   0.01 $\pm$ 0.00 &  0.23 $\pm$ 0.04 &  0.03 $\pm$ 0.02 &   0.16 $\pm$ 0.05 &    1.00 $\pm$ 0.00 \\
\cline{1-8}
\multirow{2}{*}{magic} & max RF1-GNB &  0.69 $\pm$ 0.02 &  0.58 $\pm$ 0.07 &  0.85 $\pm$ 0.02 &  0.71 $\pm$ 0.05 &   0.25 $\pm$ 0.01 &    1.00 $\pm$ 0.00 \\
          & min MLP2-DT2 &  0.58 $\pm$ 0.03 &  0.08 $\pm$ 0.02 &  0.61 $\pm$ 0.03 &  0.17 $\pm$ 0.01 &   0.11 $\pm$ 0.01 &    1.00 $\pm$ 0.00 \\
\cline{1-8}
\multirow{2}{*}{mushroom} & max KN1-GNB &  0.78 $\pm$ 0.04 &  0.84 $\pm$ 0.08 &  0.97 $\pm$ 0.02 &  0.96 $\pm$ 0.03 &   0.89 $\pm$ 0.06 &  0.67 $\pm$ 0.08 \\
          & min RF2-GNB &  0.77 $\pm$ 0.03 &  0.87 $\pm$ 0.03 &  0.98 $\pm$ 0.02 &  0.95 $\pm$ 0.04 &   0.89 $\pm$ 0.06 &  0.64 $\pm$ 0.08 \\
\cline{1-8}
\multirow{2}{*}{redwine} & max RF1-KN2 &   0.50 $\pm$ 0.12 &  0.11 $\pm$ 0.06 &   0.60 $\pm$ 0.06 &   0.40 $\pm$ 0.03 &     --- &    --- \\
          & min KN1-GNB &  0.65 $\pm$ 0.04 &  0.65 $\pm$ 0.13 &  0.74 $\pm$ 0.03 &  0.75 $\pm$ 0.03 &     --- &    --- \\
\cline{1-8}
\multirow{2}{*}{tictactoe} & max LR-GNB &  0.75 $\pm$ 0.03 &   0.80 $\pm$ 0.05 &  0.81 $\pm$ 0.02 &  0.83 $\pm$ 0.03 &    0.90 $\pm$ 0.02 &  0.77 $\pm$ 0.07 \\
          & min DT2-KN2 &    0.00 $\pm$ 0.00 &    0.00 $\pm$ 0.00 &   0.23 $\pm$ 0.20 &  0.06 $\pm$ 0.06 &     0.00 $\pm$ 0.00 &    0.00 $\pm$ 0.00 \\
\cline{1-8}
\multirow{2}{*}{waveform} & max LR-DT1 &  0.31 $\pm$ 0.03 &   0.10 $\pm$ 0.05 &  0.49 $\pm$ 0.06 &  0.08 $\pm$ 0.02 &     --- &    --- \\
          & min MLP1-RF2 &  0.21 $\pm$ 0.05 &  0.04 $\pm$ 0.01 &  0.25 $\pm$ 0.07 &  0.03 $\pm$ 0.01 &     --- &    --- \\
\cline{1-8}
\multirow{2}{*}{whitewine} & max RF1-GNB &  0.65 $\pm$ 0.03 &   0.60 $\pm$ 0.07 &  0.72 $\pm$ 0.02 &  0.77 $\pm$ 0.05 &     --- &    --- \\
          & min LR-KN2 &  0.57 $\pm$ 0.02 &  0.36 $\pm$ 0.09 &  0.69 $\pm$ 0.02 &   0.60 $\pm$ 0.02 &     --- &    --- \\
\bottomrule
\end{tabular}

\label{tab:approaches-prre-std}
\end{table*}

\begin{table*}[h]
% \small
\centering

\caption{Interpretability metrics for Direct DT and BRCG Diff. with standard deviation over 5 runs.}
\begin{tabular}{llcccc}
\toprule
          &            & \multicolumn{2}{c}{\textbf{Direct DT}} & \multicolumn{2}{c}{\textbf{BRCG Diff.}} \\
          
          \cmidrule(r){3-4} \cmidrule(l){5-6}

\textbf{Dataset} & $M_1$ \textbf{vs.} $M_2$ &            \#r &            \#p &              \#r &            \#p \\
\midrule
\multirow{2}{*}{adult} & max MLP1-GB &  18.2 $\pm$ 1.17 &  49.0 $\pm$ 2.53 &       2.0 $\pm$ 0.0 &    9.0 $\pm$ 0.71 \\
          & min MLP2-DT2 &   12.8 $\pm$ 0.4 &  45.6 $\pm$ 0.49 &       4.5 $\pm$ 0.5 &   20.0 $\pm$ 1.87 \\
\cline{1-6}
\multirow{2}{*}{bankm} & max MLP2-GB &  17.4 $\pm$ 1.02 &  54.8 $\pm$ 3.19 &      22.0 $\pm$ 7.0 &  30.25 $\pm$ 2.28 \\
          & min MLP1-GNB &  21.8 $\pm$ 1.94 &   60.6 $\pm$ 5.0 &      18.0 $\pm$ 3.0 &   24.5 $\pm$ 1.66 \\
\cline{1-6}
\multirow{2}{*}{banknote} & max KN1-GNB &   6.2 $\pm$ 1.17 &  19.4 $\pm$ 3.93 &     19.5 $\pm$ 2.29 &   27.0 $\pm$ 1.22 \\
          & min LR-DT1 &   4.8 $\pm$ 1.33 &  16.4 $\pm$ 3.72 &       6.5 $\pm$ 1.5 &  13.75 $\pm$ 2.38 \\
\cline{1-6}
\multirow{2}{*}{bc} & max DT1-GNB &    4.0 $\pm$ 1.1 &  12.8 $\pm$ 2.99 &      43.0 $\pm$ 6.6 &   31.0 $\pm$ 2.12 \\
          & min KN2-RF2 &    4.2 $\pm$ 0.4 &  13.0 $\pm$ 0.89 &    33.0 $\pm$ 10.25 &  30.25 $\pm$ 5.89 \\
\cline{1-6}
\multirow{2}{*}{diabetes} & max MLP2-GB &  11.0 $\pm$ 0.63 &  37.2 $\pm$ 2.32 &   95.25 $\pm$ 16.13 &   69.5 $\pm$ 5.59 \\
          & min RF1-GNB &   8.8 $\pm$ 3.71 &  29.2 $\pm$ 9.81 &   95.75 $\pm$ 32.86 &  61.75 $\pm$ 5.72 \\
\cline{1-6}
\multirow{2}{*}{eye} & max RF1-GNB &  19.6 $\pm$ 2.15 &  52.6 $\pm$ 5.54 &       --- &     --- \\
          & min LR-MLP1 &  23.2 $\pm$ 2.48 &   67.6 $\pm$ 4.5 &       --- &     --- \\
\cline{1-6}
\multirow{2}{*}{heloc} & max KN1-RF2 &  10.2 $\pm$ 1.94 &  35.8 $\pm$ 4.92 &      4.5 $\pm$ 1.12 &    12.5 $\pm$ 3.2 \\
          & min GB-RF1 &   8.8 $\pm$ 3.06 &  34.4 $\pm$ 9.83 &     1.75 $\pm$ 0.43 &     7.0 $\pm$ 0.0 \\
\cline{1-6}
\multirow{2}{*}{magic} & max RF1-GNB &   21.4 $\pm$ 0.8 &  65.8 $\pm$ 3.31 &    22.25 $\pm$ 2.05 &  24.25 $\pm$ 0.43 \\
          & min MLP2-DT2 &  11.2 $\pm$ 3.19 &  40.8 $\pm$ 9.26 &    14.25 $\pm$ 1.79 &   21.0 $\pm$ 3.08 \\
\cline{1-6}
\multirow{2}{*}{mushroom} & max KN1-GNB &    3.4 $\pm$ 0.8 &  12.4 $\pm$ 2.33 &    32.75 $\pm$ 1.79 &  18.75 $\pm$ 0.43 \\
          & min RF2-GNB &    3.0 $\pm$ 0.0 &  10.6 $\pm$ 0.49 &    22.75 $\pm$ 1.79 &    16.0 $\pm$ 0.0 \\
\cline{1-6}
\multirow{2}{*}{redwine} & max RF1-KN2 &  11.6 $\pm$ 2.06 &  34.4 $\pm$ 6.44 &       --- &     --- \\
          & min KN1-GNB &  18.8 $\pm$ 1.72 &  51.2 $\pm$ 5.64 &       --- &     --- \\
\cline{1-6}
\multirow{2}{*}{tictactoe} & max LR-GNB &  20.2 $\pm$ 2.14 &  34.4 $\pm$ 2.65 &  161.25 $\pm$ 12.19 &   34.5 $\pm$ 1.12 \\
          & min DT2-KN2 &    1.6 $\pm$ 0.8 &    8.4 $\pm$ 3.2 &     79.5 $\pm$ 2.96 &  30.25 $\pm$ 2.28 \\
\cline{1-6}
\multirow{2}{*}{waveform} & max LR-DT1 &  14.8 $\pm$ 1.33 &  51.0 $\pm$ 3.52 &       --- &     --- \\
          & min MLP1-RF2 &  11.2 $\pm$ 0.75 &  39.6 $\pm$ 2.15 &       --- &     --- \\
\cline{1-6}
\multirow{2}{*}{whitewine} & max RF1-GNB &   21.0 $\pm$ 2.1 &  55.2 $\pm$ 4.53 &       --- &     --- \\
          & min LR-KN2 &  24.8 $\pm$ 1.94 &  68.8 $\pm$ 3.76 &       --- &     --- \\
\bottomrule
\end{tabular}

\label{tab:approaches-interp}
\end{table*}

% References
% \clearpage % start references from a new page
% \balance % balance the 2 column references last page
% \bibliography{haldar_679}

%%%%%%%%%%%%%%%%%%%%%%%%%%%%%%%%%%%%%%%%%%
%%%%%%%%%%%%%%%%%%%%%%%%%%%%%%%%%%%%%%%%%%
%%%%%%%%%%%%%%%%%%%%%%%%%%%%%%%%%%%%%%%%%%
%%%%%%%%%%%%%%%%%%%%%%%%%%%%%%%%%%%%%%%%%%
%%%%%%%%%%%%%%%%%%%%%%%%%%%%%%%%%%%%%%%%%%
%%%%%%%%%%%%%%%%%%%%%%%%%%%%%%%%%%%%%%%%%%

\end{document}